\documentclass[10pt,twocolumn,letterpaper]{article}

 \usepackage[pagenumbers]{iccv} 

\definecolor{iccvblue}{rgb}{0.21,0.49,0.74}
\usepackage[pagebackref,breaklinks,colorlinks,allcolors=iccvblue]{hyperref}

\usepackage[ruled, linesnumbered]{algorithm2e}
\usepackage{booktabs}
\usepackage{colortbl}
\usepackage{booktabs}
\usepackage{multirow}
\usepackage{amssymb}
\usepackage{bbding}
\usepackage{bm} 
\usepackage{graphicx}

\definecolor{lightred}{rgb}{1.0, 0.78, 0.81}
\definecolor{lightpurple}{rgb}{0.88, 0.77, 1}

\newcommand{\Rmnum}[1]{\uppercase\expandafter{\romannumeral #1}}

\def\paperID{8051} 
\def\confName{ICCV}
\def\confYear{2025}

\title{DSPFusion: Image Fusion via Degradation and Semantic Dual-Prior Guidance}

\author{
	Linfeng Tang\textsuperscript{1}  \quad
	Chunyu Li\textsuperscript{1}\quad
	Guoqing Wang\textsuperscript{2} \quad
	Yixuan Yuan\textsuperscript{3} \quad
	Jiayi Ma\textsuperscript{1}\thanks{Corresponding author.} \\
	\textsuperscript{1}Wuhan University \quad
	\textsuperscript{2}University of Electronic Science and Technology of China \\
	\textsuperscript{3}The Chinese University of Hong Kong \\
	\texttt{linfeng0419@gmail.com} \quad
	\texttt{licy0089@gmail.com} \quad
	\texttt{gqwang0420@hotmail.com} \\
	\texttt{yxyuan@ee.cuhk.edu.hk} \quad
	\texttt{jyma2010@gmail.com}
}
\begin{document}
\maketitle
\begin{abstract}
Existing fusion methods are tailored for high-quality images but struggle with degraded images captured under harsh circumstances, thus limiting the practical potential of image fusion. This work presents a \textbf{D}egradation and \textbf{S}emantic \textbf{P}rior dual-guided framework for degraded image \textbf{Fusion} (\textbf{DSPFusion}), utilizing degradation priors and high-quality scene semantic priors restored via diffusion models to guide both information recovery and fusion in a unified model. In specific, it first individually extracts modality-specific degradation priors, while jointly capturing comprehensive low-quality semantic priors. Subsequently, a diffusion model is developed to iteratively restore high-quality semantic priors in a compact latent space, enabling our method to be over $20 \times$ faster than mainstream diffusion model-based image fusion schemes. Finally, the degradation priors and high-quality semantic priors are employed to guide information enhancement and aggregation via the dual-prior guidance and prior-guided fusion modules. Extensive experiments demonstrate that DSPFusion mitigates most typical degradations while integrating complementary context with minimal computational cost, greatly broadening the application scope of image fusion.
\end{abstract}

\section{Introduction}
Image fusion is a fundamental enhancement technique designed to combine complementary context from multiple images, thereby overcoming limitations of single-modality or single-type sensors~\citep{Zhang2021survey}. Infrared-visible image fusion (IVIF) is a key research area in image fusion, integrating essential thermal information from infrared (IR) images with the rich textures of visible (VI) images for comprehensive scene characterization~\citep{Zhang2023Survey}. The complete information integration and visually pleasing results make IVIF widely applied in military detection~\citep{Muller2009Military}, security surveillance~\citep{Zhang2018Surveillance}, assisted driving~\citep{Bao2023Driving}, scene understanding~\citep{Jain2023Detection, Zhang2023CMX}, \emph{etc.}

Recently, IVIF has garnered significant attention, leading to rapid advancements in relevant algorithms, which can be classified into convolutional neural network-~\citep{Ma2021STDFusionNet, Zhao2023MetaFusion}, autoencoder-~\citep{Li2018DenseFuse, Li2023LRRNet}, generative adversarial network-~\citep{Ma2019FusionGAN, Liu2022TarDAL}, Transformer-~\citep{Ma2022SwinFusion, Zhang2022Transformer}, and diffusion model (DM)-based~\citep{Zhao2023DDFM, Yi2024Diff-IF} methods. Moreover, from a functional perspective, they can be categorized into visual-oriented~\citep{Ma2019FusionGAN, Tang2022PIAFusion}, degradation-aware~\citep{Tang2023DIVFusion, Zhang2024DDBF}, semantic-driven~\citep{Tang2022SeAFusion, Liu2023SegMiF}, and joint registration-fusion~\citep{Tang2022SuperFusion, Xu2023MURF} schemes. Despite the satisfactory fusion performance achieved by these methods, several challenges still remain. On the one hand, while DMs with powerful generative abilities could bring gains, DM-based fusion methods~\citep{Zhao2023DDFM, Yue2023Dif-fusion} are often computationally intensive and time-consuming, making them inapplicable in real-time tasks. On the other hand, although some degradation-aware methods have been proposed to address imaging interferences, they still struggle with complex fusion scenarios. For example, DIVFusion~\citep{Tang2023DIVFusion} and PAIF~\citep{Liu2023PAIF} are tailored for specific degradations (\emph{e.g.,} low-light or noise) yet fail to generalize to others. Additionally, some general degradation-aware methods handle multiple degradations within a single framework assisted by additional semantic context, such as text prompts~\citep{Yi2024Text-IF}. However, they are sensitive to text prompts and struggle to handle cases where degradations occur simultaneously in both infrared and visible images. Moreover, tailoring text descriptions for each fusion scenario is challenging.

%

To overcome the above challenges, we propose a degradation and semantic prior dual-guided image fusion framework, abbreviated as DSPFusion, which incorporates degradation suppression and information aggregation in a unified model without additional assistance. The proposed method involves two training phases. In Stage~\Rmnum{1}, the semantic prior embedding network (SPEN) captures the semantic prior from cascaded high-quality sources, while the degradation prior embedding network (DPEN) extracts distinct degradation priors from two degraded images separately. A Transformer-based restoration and fusion network, guided by degradation and semantic priors, synthesizes high-quality fusion results. Note that the scene semantic prior is jointly derived from both modalities, enabling our model to enhance one using high-quality context from its complementary modality. We employ a contrastive mechanism to train DPEN, thus ensuring degradation priors effectively characterize various degradation types. Since high-quality source images are unavailable in practical situations, we deploy a diffusion model to restore high-quality semantic priors from low-quality ones in Stage~\Rmnum{2}. The diffusion process is performed in a compact latent space, making our model computationally efficient and lightweight. Ultimately, DPEN adaptively identifies degradation types and a diffusion model refines semantic priors, aiding the restoration and fusion model in synthesizing high-quality fusion results. In summary, our main contributions are as follows:
\begin{itemize}
	\item[-] We propose a novel restoration and fusion framework with dual guidance of degradation and semantic priors, effectively handling most typical degradations (\emph{e.g.}, low-light, over-exposure, noise, blur, and low-contrast) while aggregating complementary information from multiple modalities in a unified model. To our knowledge, it is the first model that comprehensively addresses various degradations in image fusion.
	\item[-] A diffusion model is devised to restore high-quality semantic priors in a compact latent space, providing coarse-grained semantic guidance with low computational cost.
	\item[-] A contrastive mechanism is employed to constrain DPEN to adaptively perceive degradation types, thereby guiding the restoration and fusion network as well as the semantic prior diffusion model to purposefully handle degradations without requiring additional auxiliary information.
	\item[-] Extensive experiments on normal and degraded scenarios demonstrate the superiority of our method in degradation suppression and complementary context aggregation.
\end{itemize}

\section{Related Work} \label{sec:related}
\textbf{Image Fusion.}
Earlier visual-oriented fusion methods focus on integrating cross-modal complementary context and enhancing visual quality, which rely on elaborate network architectures and loss functions to preserve complementary information that remains faithful to source images. Initially, mainstream network architectures primarily include CNN~\citep{Liang2022DeFusion, Zhao2023MetaFusion}, AE~\citep{Li2023LRRNet, Zhao2023CDDFuse}, and GAN~\citep{Ma2019FusionGAN, Liu2022TarDAL}. With the rise of Transformers~\citep{Vaswani2017Attention} and diffusion models~\citep{Ho2020DDPM}, these architectures gradually dominate fusion model design~\citep{Ma2022SwinFusion, Yue2023Dif-fusion}. However, as a compute-intensive process, the time cost of diffusion models remains a contentious issue.

Furthermore, several schemes including joint registration and fusion~\citep{Xu2022RFNet, Wang2022UMF, Xu2023MURF}, semantic-driven~\citep{Tang2022SeAFusion, Liu2022TarDAL, Sun2022Detfusion}, and degradation-aware~\citep{Tang2023DIVFusion, Liu2023PAIF} methods, are proposed to broaden the practical applications of image fusion. Particularly, under some extreme conditions, environmental factors like low light and noise inevitably affect imaging. Thus,~\citet{Tang2023DIVFusion} proposed an illumination-robust fusion method, achieving low-light enhancement and complementary context aggregation simultaneously. \citet{Liu2023PAIF} developed a perception-aware method by leveraging adversarial attack and architecture search to boost the robustness against noise. However, these methods are tailored to specific degradations and struggle with complex and diverse interferences. To this end, \citet{Yi2024Text-IF} leveraged CLIP to extract semantic embedding from texts to assist the fusion network in addressing multiple degradations. However, customizing text for each scenario is costly and impractical. Therefore, it is urgent and challenging to directly identify degradation types from source images, allowing a unified network to effectively handle diverse degradations and achieve optimal information aggregation.

\textbf{Unified Image Restoration.}
With advancements in deep learning technology, the field of image restoration is evolving beyond designing specialized models for specific degradation factors. Initially, researchers modeled different degradations uniformly and trained task-specific headers~\citep{Chen2021IPT} or separate models~\citep{Zamir2022Restormer, Xia2023DiffIR} to address various degradations. Furthermore, PromptIR~\citep{Potlapalli2024PromptIR} and AutoDIR~\citep{Jiang2024AutoDIR} interpret textual user requirements via the CLIP encoder~\citep{Radford2021CLIP}, guiding the general restoration models to deal with diverse degradations. To avoid reliance on user input,~\cite{Li2022AirNet} employed contrastive learning to identify degradation types from corrupted images and guide restoration models in addressing corruptions via feature modulation. Similarly, \cite{Luo2024DA-CLIP} fine-tuned CLIP on their mixed degradation dataset to develop DA-CLIP, which directly perceives degradation types and predicts high-quality content embeddings from corrupted inputs, aiding restoration networks in handling various degradations. Note that these unified restoration models are designed for natural images and usually not applicable to multi-modal images, such as infrared and visible images.

\begin{figure*}[t]
	\centering
	\includegraphics[width=0.965\linewidth]{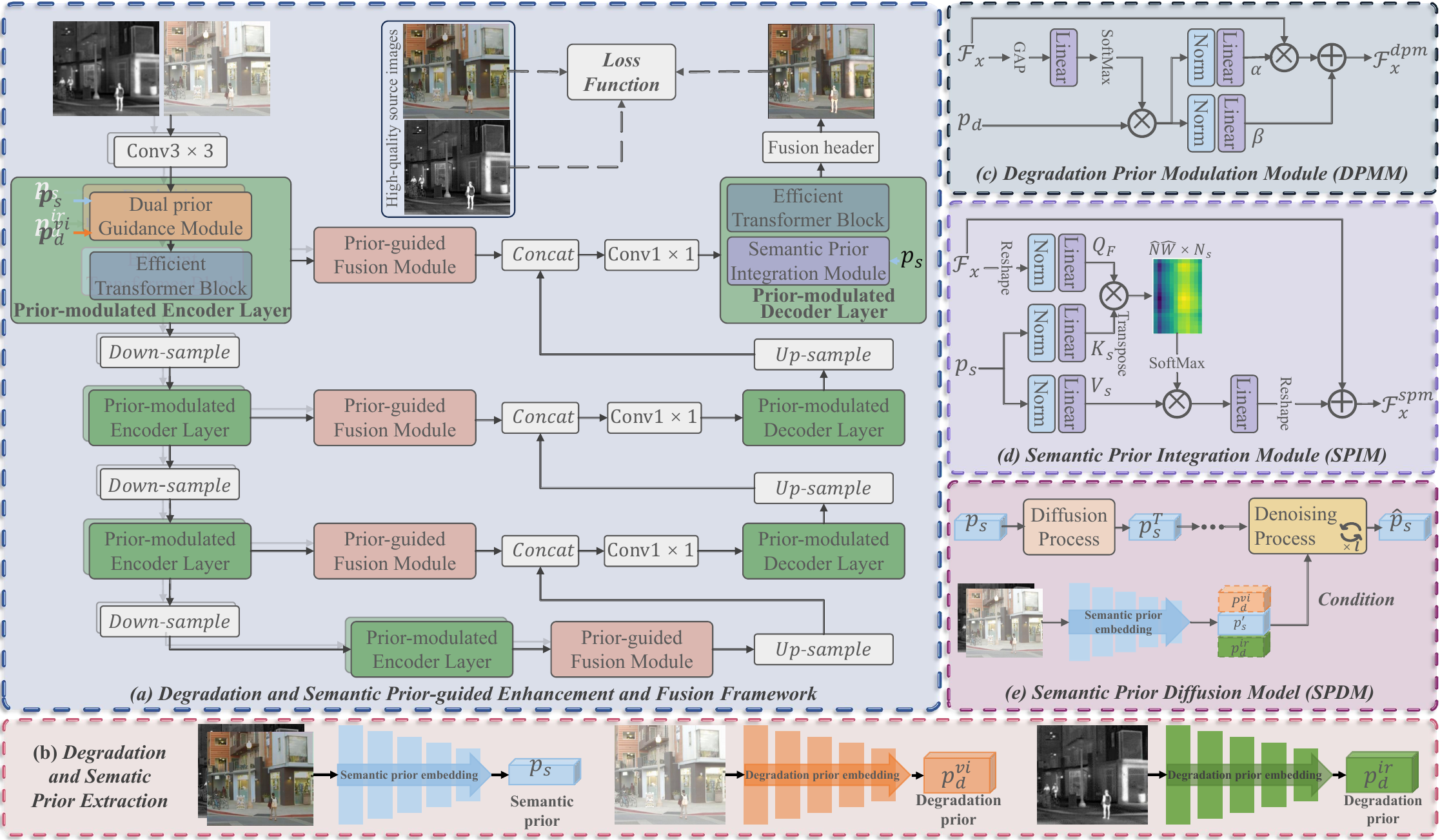}	
	\vspace{-0.05in}
	\caption{The overall framework of our image fusion network with degradation and semantic dual-prior guidance.}
	\label{fig:Framework}
\end{figure*}

\textbf{Diffusion Model.}
Benefiting from their powerful generative capabilities, diffusion models~(DMs) have been applied to diverse applications such as text-to-image generation~\citep{Rombach2022LDM}, image restoration~\citep{Xia2023DiffIR}, super-resolution~\citep{Saharia2022SR3}, deblurring~\citep{Chen2024Hi-Diff}, and deraining~\citep{Ozdenizci2023WeatherDiff}, consistently delivering impressive results. DMs have also been applied to the image fusion task. \citet{Yue2023Dif-fusion} utilized the denoising network of DMs to enhance feature extraction. \citet{Zhao2023DDFM} integrated a pre-trained DM into the EM algorithm, achieving multi-modal fusion with generative priors from natural images. However, these schemes perform the diffusion process in the image domain, making DM-based methods time-consuming. To improve efficiency, some approaches, such as Stable Diffusion~\citep{Rombach2022LDM}, PVDM~\citep{Yu2023PVDM}, and Hi-Diff~\citep{Chen2024Hi-Diff}, transfer the diffusion process into a latent space. 

\section{Methodology}
\subsection{Overview}
Our workflow is illustrated in Fig.~\ref{fig:Framework}. Given low-quality visible image $I_{vi}^{lq}$ and infrared image $I_{ir}^{lq}$, we first extract global low-quality semantic prior ($\hat{p}_s$) and degradation priors ($p_d^{vi}$ and $p_d^{ir}$) via the semantic prior embedding network (SPEN, $\bm{\mathcal{N}}_s$) and degradation prior embedding network (DPEN, $\bm{\mathcal{N}}_d$), described as:
\begin{equation}
	\begin{aligned}
	\hat{p}_s &= \bm{\mathcal{N}}_s(I_{vi}^{lq}, I_{ir}^{lq}; \theta_s), \\
	\{p_d^{vi}, p_d^{ir}\} &= \{\bm{\mathcal{N}}_d(I_{vi}^{lq}; \theta_d), \bm{\mathcal{N}}_d(I_{ir}^{lq}; \theta_d) \}, \label{eq:priors}
	\end{aligned}
\end{equation}
where $\theta$ denotes the network parameter. Then, a semantic prior diffusion model (SPDM, $\bm{\mathcal{N}}_{dm}$) is designed to restore high-quality semantic prior~($p_s^\prime$) from $\hat{p}_s$ guided by degradation priors, which is formulated as:
\begin{equation}
	p^\prime_s = \bm{\mathcal{N}}_{dm}(\hat{p}_s, p_d^{vi}, p_d^{ir}; \theta_{dm}). \label{eq:semantic_prior_diffusion}
\end{equation}
Finally, $p_s^\prime$, $p_d^{vi}$, and $p_d^{ir}$ are employed together to assist the restoration and fusion network ($\bm{\mathcal{N}}_{ef}$) in synthesizing high-quality fused images ($I_f$):
\begin{equation}
	I_f = \bm{\mathcal{N}}_{ef}(I_{vi}^{lq}, I_{ir}^{lq}, p_d^{vi}, p_d^{ir}, p_s^\prime; \theta_{ef}). \label{eq:fusion}
\end{equation}
$\bm{\mathcal{N}}_{ef}$ is a successor of Restormer~\citep{Zamir2022Restormer}. Specifically, we develop two parallel branches to extract multi-scale visible and infrared features, while integrating the degradation and semantic priors to counteract various degradations. The $k$-th level feature extraction is defined as $\mathcal{F}_{x}^k = E_k(F_{x}^{k-1}, p_d^{x}, p_s)$, where $x\in\{ir, vi\}$, $E_k$ denotes the $k$-th level prior-modulated encoder layer. Then, the semantic prior-guided fusion module~(PGFM, $\bm{\mathcal{M}}_f$) is employed to aggregate the complementary information on each level and output $F_f$, formulated as $ \mathcal{F}_f^k = \bm{\mathcal{M}}_f(\mathcal{F}_{ir}^k, \mathcal{F}_{vi}^k, p_s)$.
The prior-modulated decoder layers then refine fused features from coarse to fine-grained. Finally, a fusion header generates high-quality fusion results ($I_f$). Following previous practice~\citep{Xia2023DiffIR, Chen2024Hi-Diff}, we train our DSPFusion with a two-stage training strategy, where Stage~\Rmnum{1} focuses on prior extraction and modulation, and Stage~\Rmnum{2} optimizes the SPDM. 

\subsection{Stage~\Rmnum{1}: Prior Extraction and Modulation}
Stage~\Rmnum{1} aims to compress high- and low-quality images into a compact latent space to characterize scene semantics and degradation types, thus guiding restoration and fusion.

\subsubsection{Network Architectures}
\textbf{Degradation and Semantic Embedding.} As shown in Fig.~\ref{fig:Framework}~(b), high-quality images~$I_{vi}^{hq}$ and $I_{ir}^{hq}$ are jointly fed into the semantic prior embedding network (SPEN) to obtain a compact semantic prior~$p_s$. In parallel,  $I_{vi}^{lq}$ and $I_{ir}^{lq}$ are processed separately by the degradation prior embedding network (DPEN) to capture degradation priors~$p_d^{vi}$ and $p_d^{ir}$. SPEN and DPEN share a similar structure with residual blocks to generate prior embeddings~$p \in \mathbb{R}^{N \times C'}$, where $N$ and $C'$ represent the token number and channel dimension. Notably, $N$ is much smaller than $H \times W$, resulting in a higher compression ratio ($\frac{H \times W}{N_s}$) compared to previous latent diffusion models (\emph{e.g.}, $8$)~\cite{Rombach2022LDM}, significantly reducing the computational burden of subsequent SPDM. Additionally, the distribution of the latent semantic space~($\mathbb{R}^{N \times C'}$) is simpler than that of the image space~($\mathbb{R}^{H \times W \times 3}$), which can be approximated with fewer iterations. Thus, SPDM only requires fewer sampling steps ($T\ll1000$) to infer semantic priors compared to mainstream image-level DM-based fusion schemes~\citep{Zhao2023DDFM}, further decreasing computational cost.

\textbf{Dual Prior Guidance Module.} We integrate these priors into $\bm{\mathcal{N}}_{ef}$ via the dual prior guidance module~(DPGM). In particular, input features $\mathcal{F}_x$ first pass through the parallel degradation prior modulation module~(DPMM) and semantic prior integration module~(SPIM). As shown in Fig.~\ref{fig:Framework}~(c), $\mathcal{F}_x$ is compressed into a vector matching the size of $p_d^x$ and then multiplied by $p_d^x$. The resulting product passes through a linear layer to output the modulation parameters $\alpha_d^x$ and $\beta_d^x$. Then, DPMM is formulated as $\mathcal{F}_x^{dpm} = (\alpha_d^x \otimes \mathcal{F}_x) \oplus  \beta_d^x$, referring to~\citep{Li2022AirNet, Yi2024Text-IF}.
In parallel, the semantic prior is integrated into $\mathcal{F}_x$ through SPIM to enhance its global perception of high-quality scene context. As shown in Fig.~\ref{fig:Framework}~(d), $\mathcal{F}_x$ is mapped as a query $Q_F\in\mathbb{R}^{\hat{H}\hat{W}\times C'}$, and $p_s$ is mapped as the key $K_s\in \mathbb{R}^{N_s\times C'}$ and value $V_s \in \mathbb{R}^{N_s\times C'}$. Then, cross-attention is applied to perform semantic prior integration and generate the semantic-modulated features as:
\begin{equation}
	\mathcal{F}_x^{spi} = \mathcal{F}_x \oplus \text{softmax}\left(Q_FK_s^T / \sqrt{d_k}\right)V_s, \label{eq:SPIM}
\end{equation}
where $d_k$ is a learnable scaling factor. Then, we also employ the cross-attention mechanism to aggregate $\mathcal{F}_x^{dpm}$ and $\mathcal{F}_x^{spi}$ to obtain final reinforcement features with $\mathcal{F}_x^{dpg} = \mathcal{F}_x^{spi} \oplus \text{softmax}\left(Q_{spi}K_{dpm}^T/\sqrt{d_k}\right)V_{dpm}$, where $Q_{spi}$ is mapped from $\mathcal{F}_x^{spi}$, and $K_{dpm}$ and $V_{dpm}$ are mapped from $\mathcal{F}_x^{dpm}$. Importantly, $p_s$ provides global semantic guidance, and $p_d$ explicitly indicates degradation types, thereby reducing the overall training difficulty of restoration. 

\textbf{Prior-Guided Fusion Module.} Considering that $p_s$ is jointly extracted from multi-modal inputs, implicitly achieving complementary context aggregation, we utilize it to generate channel-wise fusion weights. Moreover, a spatial attention mechanism is utilized to calculate spatial-wise weights that, along with channel-wise weights, yield the final fusion weights $w^k_{ir}$ and $w^k_{vi}$. Finally, complementary feature fusion is formulated as: $\mathcal{F}_f^{k} = w^k_{ir} \mathcal{F}_{ir}^k \oplus w^k_{vi} \mathcal{F}_{vi}^k$.

\textbf{Image Reconstruction.} The multi-scale fused features are refined from coarse to fine using prior-modulated decoder layers, which utilize the SPIM rather than the DPGM, relying exclusively on high-quality scene semantic priors to enhance feature reinforcement. Subsequently, a fusion header, structurally similar to the decoder layer, generates $I_f$ from enhanced fused features ($\mathcal{F}_f^0$). More network details can be found in the \textbf{Supple. Material}.

\subsubsection{Loss Functions}\label{Sec:loss}
Since degradation and semantic priors are abstract high-dimensional features without ground-truth constraints, we use fusion and contrastive losses to jointly optimize $\bm{\mathcal{N}}_{ef}$, $\bm{\mathcal{N}}_{s}$, and $\bm{\mathcal{N}}_{d}$. Following~\citet{Ma2022SwinFusion} and \citet{Yi2024Text-IF}, the fusion loss involves content, structural similarity~(SSIM), and color consistency losses. To counteract degradations, we construct these losses with manually obtained high-quality source images. The content loss is defined as:
\begin{equation}
	\begin{aligned}
		\mathcal{L}_{cont} &= \frac{1}{HW} ( \| I_f - \text{max}(I_{vi}^{hq}, I_{ir}^{hq}) \|_1 \\
		&+ \gamma \cdot \| \nabla I_f - \text{max}(\nabla I_{vi}^{hq}, \nabla I_{ir}^{hq}) \|_1 ),
	\end{aligned}
\end{equation}
where $\nabla$ means the Sobel operator, $\text{max}(\cdot)$ is the maximum selection for preserving salient targets and textures, $\|\cdot\|_1$ and $\gamma$ are the $l_1$-norm and trade-off parameter. The SSIM loss is applied to maintain the structural similarity between the fused image and high-quality sources, formulated as:
\begin{equation}
	\mathcal{L}_{ssim} = ( 1 - \text{SSIM}(I_f, I_{vi}^{hq})) + ( 1 - \text{SSIM}(I_f, I_{ir}^{hq})).
\end{equation}
Moreover, we construct the color consistency loss to encourage fused images to preserve color information from high-quality visible images. It is defined as:
\begin{equation}
	\mathcal{L}_{color} = \frac{1}{HW} \| \Phi_{CbCr}(I_f) - \Phi_{CbCr}(I_{vi}^{hq}) \| _1,
\end{equation}
where $\Phi_{CbCr}(\cdot)$ converts RGB to CbCr. Besides, our DPEN aims to adaptively identify various degradations. For inputs with different degradations, the corresponding $p_d$ should be distinct, even if image contents are the same. To achieve this, we devise a contrastive loss~$\mathcal{L}_{cl}$ that pulls together priors characterizing same degradations while pushing apart priors representing various degradations. For a degradation prior~$p_d$, $q_k^+$ and $q_m^-$ are the corresponding positive and negative counterparts. Then, $\mathcal{L}_{cl}$ is formulated as:
\begin{equation}
	\mathcal{L}_{cl} = \sum_{k=1}^{K}-\log\frac{\exp(p_d\cdot q_k^+ /\tau)}{\sum_{m}^{M}\exp(p_d \cdot q_m^- / \tau)}, \label{eq:cl}
\end{equation}
where $K$ and $M$ are the number of positive and negative samples, and $\tau$ is a temperature parameter. If $p_d$ is extracted from an image with a specific degradation, then $q_k^+$ is extracted from other scenes with the same degradation, while $q_m^-$ is extracted from the same scene but with various degradations or modalities. Finally, the total loss of Stage~\Rmnum{1} is formulated as the weighted sum of aforementioned losses:
\begin{equation}
	\mathcal{L}_{I} = \lambda_{cont} \cdot \mathcal{L}_{cont} + \lambda_{ssim} \cdot \mathcal{L}_{ssim} + \lambda_{color} \cdot \mathcal{L}_{color} +
	\lambda_{cl} \cdot \mathcal{L}_{cl}, \label{eq:stageI}
\end{equation}
where $\lambda_{cont}$, $\lambda_{ssim}$, $\lambda_{color}$, and $\lambda_{cl}$ are hyper-parameters.

\begin{table*}[t]
	\centering
	\caption{Quantitative comparison results on typical fusion datasets. The best and second-best results are highlighted in \textcolor{lightred}{\textbf{Red}} and \textcolor[HTML]{7B68EE}{Purple}.} \label{tab:normal}
	\vspace{-0.1in}
	\setlength{\tabcolsep}{2pt}
	\resizebox{0.99\textwidth}{!}{
		\begin{tabular}{@{}lccccccccccccccccccccccccccccccccccc@{}}
			\toprule
			& \multicolumn{8}{l}{\textbf{MSRS}} & \textbf{} & \multicolumn{8}{l}{\textbf{LLVIP}} & \textbf{} & \multicolumn{8}{l}{\textbf{RoadScene}} & \multicolumn{1}{l}{} & \multicolumn{8}{l}{\textbf{TNO}} \\ \cmidrule(lr){2-9} \cmidrule(lr){11-18} \cmidrule(lr){20-27} \cmidrule(l){29-36}
			\multirow{-2}{*}{\textbf{Methods}} & \textbf{EN} & \textbf{} & \textbf{MI} & \textbf{} & \textbf{VIF} & \textbf{} & \textbf{Qabf} & \textbf{} & \textbf{} & \textbf{EN} &  & \textbf{MI} &  & \textbf{VIF} &  & \textbf{Qabf} & \textbf{} & \textbf{} & \textbf{EN} & \textbf{} & \textbf{MI} & \textbf{} & \textbf{VIF} & \textbf{} & \textbf{Qabf} &  &  & \textbf{EN} & \textbf{} & \textbf{MI} & \textbf{} & \textbf{VIF} & \textbf{} & \textbf{Qabf} & \textbf{} \\ \midrule
			\textbf{DeFus.}~\cite{Liang2022DeFusion} & 6.350 &  & 3.054 &  & 0.736 &  & 0.505 &  &  & 7.112 &  & 3.196 &  & 0.683 &  & 0.487 &  &  & 6.910 &  & 3.018 &  & 0.537 &  & 0.404 &  &  & 6.581 &  & 2.917 &  & 0.596 &  & 0.384 &  \\
			\textbf{PAIF}~\cite{Liu2023PAIF} & 5.830 &  & 2.907 &  & 0.470 &  & 0.329 &  &  & 6.937 &  & 2.533 &  & 0.432 &  & 0.286 &  &  & 6.750 &  & 2.919 &  & 0.387 &  & 0.248 &  &  & 6.198 &  & 2.561 &  & 0.421 &  & 0.243 &  \\
			\textbf{MetaFus.}~\cite{Zhao2023MetaFusion} & 6.355 &  & 1.693 &  & 0.700 &  & 0.476 &  &  & 6.645 &  & 1.400 &  & 0.629 &  & 0.429 &  &  & 7.363 &  & 2.195 &  & 0.517 &  & 0.416 &  &  & \cellcolor[HTML]{E6E6FA}7.184 &  & 1.815 &  & 0.615 &  & 0.362 &  \\
			\textbf{LRRNet}~\cite{Li2023LRRNet} & 6.197 &  & 2.886 &  & 0.536 &  & 0.451 &  &  & 6.006 &  & 1.749 &  & 0.397 &  & 0.281 &  &  & 7.051 &  & 2.649 &  & 0.463 &  & 0.344 &  &  & 6.944 &  & 2.577 &  & 0.577 &  & 0.352 &  \\
			\textbf{MURF}~\cite{Xu2023MURF} & 5.036 &  & 1.516 &  & 0.403 &  & 0.311 &  &  & 5.869 &  & 2.017 &  & 0.355 &  & 0.317 &  &  & 6.961 &  & 2.492 &  & 0.498 &  & 0.468 &  &  & 6.654 &  & 1.912 &  & 0.528 &  & 0.378 &  \\
			\textbf{SegMiF}~\cite{Liu2023SegMiF} & 6.109 &  & 2.472 &  & 0.774 &  & 0.565 &  &  & \cellcolor[HTML]{E6E6FA}7.172 &  & 2.819 &  & 0.837 &  & \cellcolor[HTML]{E6E6FA}0.651 &  &  & 7.254 &  & 2.657 &  & 0.615 &  & 0.543 &  &  & 6.976 &  & 3.036 &  & 0.876 &  & \cellcolor[HTML]{E6E6FA}0.589 &  \\
			\textbf{DDFM}~\cite{Zhao2023DDFM} & 6.182 &  & 2.661 &  & 0.721 &  & 0.468 &  &  & 6.814 &  & 2.590 &  & 0.632 &  & 0.475 &  &  & 7.111 &  & 2.84 &  & 0.587 &  & 0.482 &  &  & 6.878 &  & 2.408 &  & 0.691 &  & 0.466 &  \\
			\textbf{EMMA}~\cite{Zhao2024EMMA} & \cellcolor[HTML]{FFC7CE}\textbf{6.713} &  & 4.129 &  & 0.957 &  & 0.632 &  &  & 7.160 &  & \cellcolor[HTML]{E6E6FA}3.374 &  & 0.740 &  & 0.572 &  &  & \cellcolor[HTML]{FFC7CE}\textbf{7.383} &  & \cellcolor[HTML]{E6E6FA}3.140 &  & 0.605 &  & 0.461 &  &  & \cellcolor[HTML]{FFC7CE}\textbf{7.203} &  & 3.038 &  & 0.755 &  & 0.472 &  \\
			\textbf{Text-IF}~\cite{Yi2024Text-IF} & 6.648 &  & \cellcolor[HTML]{E6E6FA}4.283 &  & \cellcolor[HTML]{E6E6FA}1.031 &  & \cellcolor[HTML]{E6E6FA}0.692 &  &  & 6.961 &  & 3.142 &  & \cellcolor[HTML]{E6E6FA}0.855 &  & 0.648 &  &  & 7.299 &  & 2.988 &  & \cellcolor[HTML]{E6E6FA}0.698 &  & \cellcolor[HTML]{E6E6FA}0.588 &  &  & 7.168 &  & \cellcolor[HTML]{E6E6FA}3.524 &  & \cellcolor[HTML]{E6E6FA}0.918 &  & 0.583 &  \\
			\textbf{DSPFusion} & \cellcolor[HTML]{E6E6FA}6.695 &  & \cellcolor[HTML]{FFC7CE}\textbf{4.736} &  & \cellcolor[HTML]{FFC7CE}\textbf{1.044} &  & \cellcolor[HTML]{FFC7CE}\textbf{0.726} &  &  & \cellcolor[HTML]{FFC7CE}\textbf{7.314} &  & \cellcolor[HTML]{FFC7CE}\textbf{4.390} &  & \cellcolor[HTML]{FFC7CE}\textbf{0.943} &  & \cellcolor[HTML]{FFC7CE}\textbf{0.717} &  &  & \cellcolor[HTML]{E6E6FA}7.363 &  & \cellcolor[HTML]{FFC7CE}\textbf{3.962} &  & \cellcolor[HTML]{FFC7CE}\textbf{0.755} &  & \cellcolor[HTML]{FFC7CE}\textbf{0.66}7 &  &  & 7.152 &  & \cellcolor[HTML]{FFC7CE}\textbf{4.680} &  & \cellcolor[HTML]{FFC7CE}\textbf{0.931} &  & \cellcolor[HTML]{FFC7CE}\textbf{0.640} &  \\ \bottomrule
		\end{tabular}
	}
\end{table*}

\subsection{Stage~\Rmnum{2}: Semantic Prior Diffusion Model}
In Stage~\Rmnum{2}, we develop a semantic prior diffusion model~(SPDM) to restore high-quality semantic prior from low-quality ones, thereby guiding restoration and fusion. SPDM involves the forward diffusion and reverse denoising processes, as shown in Fig.~\ref{fig:Framework} (e). In the diffusion process, we first embed $I_{vi}^{hq}$ and $I_{ir}^{hq}$ into a high-quality semantic prior $p_s$, which is simply marked as $x_0$ in this section. $x_0$ serves as the starting point of a forward Markov chain and gradually adds Gaussian noise to it over $T$ iterations as:
\begin{equation}
	\begin{aligned}
	q(x_{1:T}|x_0) &= \Pi_{t=1}^{T} q(x_t|x_{t-1}), \\
	q(x_t|x_{t-1}) &= \mathcal{N}(x_t; \sqrt{\alpha_t}x_{t-1}, \beta_t \mathbf{I}),
	\end{aligned}
\end{equation}
where $x_t$ is the $t$-step noisy variable, $\beta_t$ governs noise variance, and $\alpha_t = 1 - \beta_t$. Through iterative derivation with reparameterization, the forward process is reformulated as:
\begin{equation}
	q(x_t|x_0) = \mathcal{N}(x_t, \sqrt{\bar{\alpha}_t}x_0, (1 - \bar{\alpha}_t)\mathbf{I}),
\end{equation}
where $\bar{\alpha}_t = \Pi_{i=1}^t \alpha_i$. As $t$ approaches a large value $T$, $\bar{\alpha}_T$ tends to 0 and $q(x_T | x_0)$ approximates the normal distribution $\mathcal{N}(0, \mathbf{I})$, thus completing the forward process.

\begin{table*}[]
\centering
\caption{Quantitative comparison results in different degraded scenarios with pre-enhancement.} \label{tab:degradation}
\vspace{-0.05in}
\setlength{\tabcolsep}{2pt}
\resizebox{0.99\textwidth}{!}{		
	\begin{tabular}{@{}lccccccccccccccccccccccccccccccccccc@{}}
		\toprule
		& \multicolumn{8}{l}{\textbf{VI (Blur)}} & \multicolumn{1}{l}{\textbf{}} & \multicolumn{8}{l}{\textbf{VI (Rain)}} & \multicolumn{1}{l}{\textbf{}} & \multicolumn{8}{l}{\textbf{VI (Low-light, LL)}} & \multicolumn{1}{l}{\textbf{}} & \multicolumn{8}{l}{\textbf{VI (Over-exposure, OE)}} \\ \cmidrule(lr){2-9} \cmidrule(lr){11-27} \cmidrule(l){29-36}
		\multirow{-2}{*}{\textbf{Methods}} & \textbf{MUSIQ} & \textbf{} & \textbf{PI} & \textbf{} & \textbf{TReS} &  & \textbf{SF} &  &  & \textbf{MUSIQ} &  & \textbf{PI} &  & \textbf{TReS} &  & \textbf{SD} &  &  & \textbf{MUSIQ} &  & \textbf{PI} &  & \textbf{TReS} &  & \textbf{SD} &  & \textbf{} & \textbf{MUSIQ} &  & \textbf{PI} &  & \textbf{TReS} &  & \textbf{SD} &  \\ \midrule
		\textbf{DeFus.}~\cite{Liang2022DeFusion} & 38.971 &  & 4.368 &  & 36.968 &  & 8.765 &  &  & 44.182 &  & 3.575 &  & 44.516 &  & 40.471 &  &  & 43.645 &  & 3.510 &  & 42.961 &  & 36.326 &  &  & 46.316 &  & 3.367 &  & 45.972 &  & 38.649 &  \\
		\textbf{PAIF}~\cite{Liu2023PAIF} & 39.363 &  & 5.431 &  & 40.298 &  & 9.102 &  &  & 45.685 &  & 4.720 &  & 46.819 &  & 38.593 &  &  & 42.432 &  & 4.725 &  & 45.248 &  & 39.239 &  &  & 37.032 &  & 5.261 &  & 35.094 &  & 52.062 &  \\
		\textbf{MetaFus.}~\cite{Zhao2023MetaFusion} & 36.674 &  & 4.946 &  & 34.451 &  & \cellcolor[HTML]{FFC7CE}\textbf{18.167} &  &  & 41.160 &  & 4.127 &  & 39.105 &  & 49.808 &  &  & 38.775 &  & 4.061 &  & 33.915 &  & \cellcolor[HTML]{E6E6FA}46.852 &  &  & 43.590 &  & 3.217 &  & 38.889 &  & 52.473 &  \\
		\textbf{LRRNet}~\cite{Li2023LRRNet} & 41.490 &  & 4.074 &  & 41.774 &  & 11.084 &  &  & 48.591 &  & 3.218 &  & 50.646 &  & 42.612 &  &  & 44.037 &  & 3.391 &  & 47.584 &  & 31.204 &  &  & 48.263 &  & 2.803 &  & 51.459 &  & 44.852 &  \\
		\textbf{MURF}~\cite{Xu2023MURF} & \cellcolor[HTML]{E6E6FA}45.860 &  & \cellcolor[HTML]{E6E6FA}3.379 &  & 46.603 &  & 12.374 &  &  & 48.703 &  & 3.131 &  & 49.965 &  & 25.117 &  &  & 44.979 &  & 3.144 &  & 46.383 &  & 21.418 &  &  & \cellcolor[HTML]{E6E6FA}52.484 &  & 2.518 &  & 56.603 &  & 33.128 &  \\
		\textbf{SegMiF}~\cite{Liu2023SegMiF} & 42.907 &  & 4.02 &  & 41.349 &  & 12.636 &  &  & 47.850 &  & 2.909 &  & 50.016 &  & 46.861 &  &  & 45.16 &  & 3.049 &  & 46.311 &  & 45.614 &  &  & 51.282 &  & 2.688 &  & 54.419 &  & 50.445 &  \\
		\textbf{DDFM}~\cite{Zhao2023DDFM} & 42.005 &  & 3.928 &  & 42.765 &  & 9.474 &  &  & 47.117 &  & 3.229 &  & 49.757 &  & 36.140 &  &  & 44.022 &  & 3.346 &  & 47.090 &  & 32.197 &  &  & 52.182 &  & 2.435 &  & \cellcolor[HTML]{FFC7CE}\textbf{59.047} &  & 41.834 &  \\
		\textbf{EMMA}~\cite{Zhao2024EMMA} & 41.442 &  & 4.128 &  & 37.81 &  & 14.036 &  &  & 48.315 &  & 3.078 &  & 45.750 &  & \cellcolor[HTML]{FFC7CE}\textbf{55.989} &  &  & 45.124 &  & 3.201 &  & 42.383 &  & 45.881 &  &  & 49.446 &  & 2.651 &  & 48.389 &  & \cellcolor[HTML]{E6E6FA}54.538 &  \\
		\textbf{Text-IF}~\cite{Yi2024Text-IF} & 44.536 &  & 3.665 &  & \cellcolor[HTML]{E6E6FA}47.524 &  & 15.153 &  &  & \cellcolor[HTML]{E6E6FA}50.109 &  & \cellcolor[HTML]{E6E6FA}2.775 &  & \cellcolor[HTML]{FFC7CE}\textbf{56.966} &  & 54.842 &  &  & \cellcolor[HTML]{E6E6FA}46.015 &  & \cellcolor[HTML]{E6E6FA}2.994 &  & \cellcolor[HTML]{E6E6FA}50.279 &  & \cellcolor[HTML]{FFC7CE}\textbf{51.537} &  &  & 52.048 &  & \cellcolor[HTML]{E6E6FA}2.290 &  & 56.979 &  & 52.200 &  \\
		\textbf{DSPFusion} & \cellcolor[HTML]{FFC7CE}\textbf{47.137} &  & \cellcolor[HTML]{FFC7CE}\textbf{2.972} &  & \cellcolor[HTML]{FFC7CE}\textbf{49.750} &  & \cellcolor[HTML]{E6E6FA}15.693 &  &  & \cellcolor[HTML]{FFC7CE}\textbf{50.467} &  & \cellcolor[HTML]{FFC7CE}\textbf{2.557} &  & \cellcolor[HTML]{E6E6FA}56.528 &  & \cellcolor[HTML]{E6E6FA}55.599 &  &  & \cellcolor[HTML]{FFC7CE}\textbf{48.500} &  & \cellcolor[HTML]{FFC7CE}\textbf{2.768} &  & \cellcolor[HTML]{FFC7CE}\textbf{54.090} &  & 45.940 &  &  & \cellcolor[HTML]{FFC7CE}\textbf{52.812} &  & \cellcolor[HTML]{FFC7CE}\textbf{2.206} &  & \cellcolor[HTML]{E6E6FA}57.198 &  & \cellcolor[HTML]{FFC7CE}\textbf{54.840} &  \\ \midrule
		& \multicolumn{8}{l}{\textbf{VI (Random noise, RN)}} & \textbf{} & \multicolumn{8}{l}{\textbf{IR (Low-contrast, LC)}} & \textbf{} & \multicolumn{8}{l}{\textbf{IR (Random noise, RN)}} & \textbf{} & \multicolumn{8}{l}{\textbf{IR (Stripe noise, SN)}} \\ \cmidrule(lr){2-9} \cmidrule(lr){11-18} \cmidrule(lr){20-27} \cmidrule(l){29-36}
		\multirow{-2}{*}{\textbf{Methods}} & \textbf{MUSIQ} & \textbf{} & \textbf{PI} & \textbf{} & \textbf{TReS} &  & \textbf{EN} &  & \textbf{} & \textbf{MUSIQ} &  & \textbf{PI} &  & \textbf{TReS} &  & \textbf{SD} &  & \textbf{} & \textbf{MUSIQ} &  & \textbf{PI} &  & \textbf{TReS} &  & \textbf{EN} &  & \textbf{} & \textbf{MUSIQ} &  & \textbf{PI} &  & \textbf{TReS} &  & \textbf{EN} &  \\ \midrule
		\textbf{DeFus.}~\cite{Liang2022DeFusion} & 34.661 &  & 4.883 &  & 30.837 &  & 7.028 &  &  & 44.378 &  & 3.538 &  & 45.154 &  & 39.753 &  &  & 40.463 &  & 3.735 &  & 43.565 &  & 7.065 &  &  & 42.603 &  & 3.593 &  & 42.966 &  & 6.968 &  \\
		\textbf{PAIF}~\cite{Liu2023PAIF} & 33.875 &  & 6.306 &  & 33.362 &  & 6.731 &  &  & 47.192 &  & 4.525 &  & 47.946 &  & 38.499 &  &  & 47.198 &  & 4.456 &  & 49.016 &  & 6.772 &  &  & 47.403 &  & 4.482 &  & 48.921 &  & 6.747 &  \\
		\textbf{MetaFus.}~\cite{Zhao2023MetaFusion} & 32.918 &  & 4.969 &  & 27.623 &  & 7.362 &  &  & 40.136 &  & 4.224 &  & 39.847 &  & 55.705 &  &  & 40.236 &  & 4.304 &  & 39.461 &  & \cellcolor[HTML]{E6E6FA}7.419 &  &  & 39.543 &  & 4.215 &  & 38.837 &  & \cellcolor[HTML]{FFC7CE}\textbf{7.448} &  \\
		\textbf{LRRNet}~\cite{Li2023LRRNet} & 34.856 &  & 4.642 &  & 32.07 &  & 7.108 &  &  & 48.540 &  & 3.189 &  & 50.456 &  & 42.697 &  &  & 46.625 &  & 3.327 &  & 49.883 &  & 7.015 &  &  & 47.382 &  & 3.248 &  & 48.708 &  & 7.016 &  \\
		\textbf{MURF}~\cite{Xu2023MURF} & \cellcolor[HTML]{E6E6FA}40.866 &  & \cellcolor[HTML]{E6E6FA}3.386 &  & \cellcolor[HTML]{E6E6FA}42.447 &  & 6.233 &  &  & 48.958 &  & 3.063 &  & 51.079 &  & 26.875 &  &  & \cellcolor[HTML]{E6E6FA}49.799 &  & 3.276 &  & 49.431 &  & 6.204 &  &  & 47.135 &  & 3.202 &  & 47.908 &  & 6.223 &  \\
		\textbf{SegMiF}~\cite{Liu2023SegMiF} & 37.362 &  & 4.081 &  & 31.640 &  & 6.975 &  &  & 48.335 &  & 2.803 &  & 51.976 &  & 50.813 &  &  & 47.909 &  & 2.977 &  & 50.489 &  & 6.997 &  &  & 46.971 &  & 2.852 &  & 49.394 &  & 7.080 &  \\
		\textbf{DDFM}~\cite{Zhao2023DDFM} & 37.325 &  & 4.351 &  & 36.870 &  & 6.940 &  &  & 48.039 &  & 3.222 &  & 51.181 &  & 37.090 &  &  & 46.829 &  & 3.512 &  & 48.653 &  & 6.960 &  &  & 45.310 &  & 3.361 &  & 47.234 &  & 6.908 &  \\
		\textbf{EMMA}~\cite{Zhao2024EMMA} & 34.754 &  & 4.432 &  & 29.364 &  & \cellcolor[HTML]{FFC7CE}\textbf{7.44} &  &  & 48.721 &  & 2.978 &  & 46.842 &  & \cellcolor[HTML]{FFC7CE}\textbf{58.299} &  &  & 45.870 &  & 3.124 &  & 45.667 &  & \cellcolor[HTML]{FFC7CE}\textbf{7.453} &  &  & 47.286 &  & 3.036 &  & 45.717 &  & 7.439 &  \\
		\textbf{Text-IF}~\cite{Yi2024Text-IF} & 39.200 &  & 3.930 &  & 36.588 &  & \cellcolor[HTML]{E6E6FA}7.406 &  &  & \cellcolor[HTML]{E6E6FA}50.022 &  & \cellcolor[HTML]{E6E6FA}2.794 &  & \cellcolor[HTML]{E6E6FA}55.203 &  & \cellcolor[HTML]{E6E6FA}56.117 &  &  & 48.795 &  & \cellcolor[HTML]{E6E6FA}2.944 &  & \cellcolor[HTML]{E6E6FA}54.003 &  & 7.402 &  &  & \cellcolor[HTML]{E6E6FA}49.376 &  & \cellcolor[HTML]{E6E6FA}2.829 &  & \cellcolor[HTML]{E6E6FA}53.717 &  & \cellcolor[HTML]{E6E6FA}7.440 &  \\
		\textbf{DSPFusion} & \cellcolor[HTML]{FFC7CE}\textbf{47.718} &  & \cellcolor[HTML]{FFC7CE}\textbf{2.954} &  & \cellcolor[HTML]{FFC7CE}\textbf{52.787} &  & 7.343 &  &  & \cellcolor[HTML]{FFC7CE}\textbf{50.597} &  & \cellcolor[HTML]{FFC7CE}\textbf{2.623} &  & \cellcolor[HTML]{FFC7CE}\textbf{56.988} &  & 56.060 &  &  & \cellcolor[HTML]{FFC7CE}\textbf{50.752} &  & \cellcolor[HTML]{FFC7CE}\textbf{2.852} &  & \cellcolor[HTML]{FFC7CE}\textbf{57.266} &  & 7.349 &  &  & \cellcolor[HTML]{FFC7CE}\textbf{51.055} &  & \cellcolor[HTML]{FFC7CE}\textbf{2.787} &  & \cellcolor[HTML]{FFC7CE}\textbf{57.243} &  & 7.353 &  \\ \midrule
		& \multicolumn{8}{l}{\textbf{VI (Rain) and IR (LC)}} & \textbf{} & \multicolumn{8}{l}{\textbf{VI (Rain) and IR (SN)}} & \textbf{} & \multicolumn{8}{l}{\textbf{VI (LL) and IR (LC)}} & \textbf{} & \multicolumn{8}{l}{\textbf{VI (LL) and IR (SN)}} \\ \cmidrule(lr){2-9} \cmidrule(lr){11-18} \cmidrule(lr){20-27} \cmidrule(l){29-36}
		\multirow{-2}{*}{\textbf{Methods}} & \textbf{MUSIQ} & \textbf{} & \textbf{PI} &  & \textbf{TReS} &  & \textbf{SD} &  & \textbf{} & \textbf{MUSIQ} &  & \textbf{PI} &  & \textbf{TReS} &  & \textbf{EN} &  & \textbf{} & \textbf{MUSIQ} &  & \textbf{PI} &  & \textbf{TReS} &  & \textbf{SD} &  &  & \textbf{MUSIQ} &  & \textbf{PI} &  & \textbf{TReS} &  & \textbf{EN} &  \\ \midrule
		\textbf{DeFus.}~\cite{Liang2022DeFusion} & 44.168 &  & 3.727 &  & 44.601 &  & 35.803 &  &  & 41.933 &  & 3.808 &  & 42.020 &  & 6.827 &  &  & 42.000 &  & 3.654 &  & 41.607 &  & 30.416 &  &  & 40.279 &  & 3.684 &  & 38.722 &  & 6.788 &  \\
		\textbf{PAIF}~\cite{Liu2023PAIF} & 46.172 &  & 4.817 &  & 46.361 &  & 37.658 &  &  & 46.013 &  & 4.778 &  & 47.135 &  & 6.661 &  &  & 42.671 &  & 4.639 &  & 46.503 &  & 30.425 &  &  & 41.686 &  & 4.796 &  & 44.084 &  & 6.616 &  \\
		\textbf{MetaFus.}~\cite{Zhao2023MetaFusion} & 40.528 &  & 4.221 &  & 39.237 &  & 51.363 &  &  & 40.175 &  & 4.214 &  & 38.272 &  & 7.339 &  &  & 38.832 &  & 3.833 &  & 33.506 &  & 43.002 &  &  & 37.957 &  & 4.086 &  & 31.719 &  & 7.194 &  \\
		\textbf{LRRNet}~\cite{Li2023LRRNet} & 48.396 &  & 3.248 &  & 50.05 &  & 42.434 &  &  & 47.151 &  & 3.335 &  & 48.290 &  & 6.999 &  &  & 43.231 &  & 3.414 &  & 45.943 &  & 31.170 &  &  & 42.730 &  & 3.564 &  & \cellcolor[HTML]{E6E6FA}44.534 &  & 6.630 &  \\
		\textbf{MURF}~\cite{Xu2023MURF} & 49.221 &  & 3.082 &  & 50.669 &  & 26.013 &  &  & 47.048 &  & 3.271 &  & 47.253 &  & 6.148 &  &  & 44.943 &  & 3.117 &  & 45.792 &  & 18.915 &  &  & 43.006 &  & 3.300 &  & 42.568 &  & 5.910 &  \\
		\textbf{SegMiF}~\cite{Liu2023SegMiF} & 48.392 &  & 2.915 &  & 50.802 &  & 46.525 &  &  & 46.548 &  & 2.998 &  & 47.923 &  & 6.989 &  &  & 44.512 &  & 2.965 &  & 45.744 &  & 40.511 &  &  & 43.337 &  & \cellcolor[HTML]{E6E6FA}3.218 &  & 43.201 &  & 6.982 &  \\
		\textbf{DDFM}~\cite{Zhao2023DDFM} & 47.513 &  & 3.318 &  & 50.074 &  & 36.381 &  &  & 44.869 &  & 3.447 &  & 46.338 &  & 6.873 &  &  & 43.242 &  & 3.473 &  & 46.691 &  & 29.616 &  &  & 41.600 &  & 3.788 &  & 43.711 &  & 6.722 &  \\
		\textbf{EMMA}~\cite{Zhao2024EMMA} & 48.864 &  & 3.017 &  & 46.307 &  & 54.659 &  &  & 47.216 &  & 3.119 &  & 45.023 &  & \cellcolor[HTML]{E6E6FA}7.366 &  &  & 44.251 &  & 3.211 &  & 41.010 &  & 43.038 &  &  & \cellcolor[HTML]{E6E6FA}43.652 &  & 3.233 &  & 39.788 &  & \cellcolor[HTML]{E6E6FA}7.209 &  \\
		\textbf{Text-IF}~\cite{Yi2024Text-IF} & \cellcolor[HTML]{E6E6FA}50.380 &  & \cellcolor[HTML]{E6E6FA}2.765 &  & \cellcolor[HTML]{FFC7CE}\textbf{56.429} &  & \cellcolor[HTML]{FFC7CE}\textbf{56.381} &  &  & \cellcolor[HTML]{E6E6FA}48.221 &  & \cellcolor[HTML]{E6E6FA}2.885 &  & \cellcolor[HTML]{E6E6FA}53.446 &  & \cellcolor[HTML]{FFC7CE}\textbf{7.400} &  &  & \cellcolor[HTML]{E6E6FA}47.815 &  & \cellcolor[HTML]{E6E6FA}2.915 &  & \cellcolor[HTML]{E6E6FA}49.334 &  & \cellcolor[HTML]{E6E6FA}45.733 &  &  & 39.866 &  & 3.272 &  & 39.043 &  & 7.095 &  \\
		\textbf{DSPFusion} & \cellcolor[HTML]{FFC7CE}\textbf{50.672} &  & \cellcolor[HTML]{FFC7CE}\textbf{2.607} &  & \cellcolor[HTML]{E6E6FA}56.362 &  & \cellcolor[HTML]{E6E6FA}55.686 &  &  & \cellcolor[HTML]{FFC7CE}\textbf{50.883} &  & \cellcolor[HTML]{FFC7CE}\textbf{2.737} &  & \cellcolor[HTML]{FFC7CE}\textbf{57.098} &  & 7.349 &  &  & \cellcolor[HTML]{FFC7CE}\textbf{48.547} &  & \cellcolor[HTML]{FFC7CE}\textbf{2.819} &  & \cellcolor[HTML]{FFC7CE}\textbf{53.892} &  & \cellcolor[HTML]{FFC7CE}\textbf{46.144} &  &  & \cellcolor[HTML]{FFC7CE}\textbf{48.711} &  & \cellcolor[HTML]{FFC7CE}\textbf{3.020} &  & \cellcolor[HTML]{FFC7CE}\textbf{54.598} &  & \cellcolor[HTML]{FFC7CE}\textbf{7.263} &  \\ \bottomrule
	\end{tabular}
}
\end{table*}

The reverse process starts from a pure Gaussian distribution and progressively denoises to generate the high-quality semantic prior via a T-step Markov chain, defined as:
\begin{equation} \label{eq:posterior}
	p(x_{t-1} | x_{t}) = \mathcal{N}(x_{t-1}; \mu(x_t, t), \sigma_t^2  \mathbf{I}),
\end{equation}
with $\mu(x_t, t) = \frac{1}{\sqrt{\alpha_t}}(x_t - \frac{\beta_t}{\sqrt{1-\bar{\alpha}_t}}\epsilon)$, and $\sigma_t^2 = \frac{(1 - \bar{\alpha}_{t-1})}{(1 - \bar{\alpha}_t)} \beta_t$.
Following previous works~\citep{Ho2020DDPM, Rombach2022LDM}, we employ a denoising U-Net ($\epsilon_{\theta}$) with the aid of the low-quality semantic prior~$\hat{p}_s$ and degradation priors~$p_d^{vi}$ and $p_d^{ir}$ to estimate the noise~$\epsilon$. Utilizing the reparameterization trick and substituting $\epsilon$ in Eq.~(\ref{eq:posterior}) with $\epsilon_{\theta}(x_t, \hat{p}_s, p_d^{vi}, p_d^{ir}, t)$, we can get:
\begin{equation}
	x_{t-1} = \frac{1}{\sqrt{\alpha_t}}\left( x_t - \frac{1 - \alpha_t}{\sqrt{1 - \bar{\alpha}_t}} \epsilon_{\theta}(x_t, \hat{p}_s, p_d^{vi}, p_d^{ir}, t)\right) + \sigma_t z.
\end{equation}

Traditionally, the objective for training $\epsilon_\theta$ is defined as:
\begin{equation}
	\nabla_\theta \| \epsilon_t - \epsilon_{\theta}(\sqrt{\bar{\alpha}_t} x_0+\sqrt{1 - \bar{\alpha}_t}\epsilon_t, \hat{p}_s, p_d^{vi}, p_d^{ir}, t)  \|^2_2.
\end{equation}
Since the distribution of the latent semantic space~($\mathbb{R}^{N_s \times C'}$) is simpler than that of the image space~($\mathbb{R}^{H \times W \times 3}$), the semantic prior~($p_s^\prime$) can be generated with fewer iterations~\citep{Chen2024Hi-Diff}. Thus, we run complete $T$ ($\ll 1000$) reverse iterations to infer $p_s^\prime$. Consequently, we use $\mathcal{L}_{diff} = \| p_s^\prime - p_s \|_1$ to train SPDM. We also apply the content, SSIM, and color consistency losses to collaboratively constrain the training of SPDM. Thus, the total loss of Stage~\Rmnum{2} is defined as:
\begin{equation}
	\mathcal{L}_{II}\! = \!\lambda_{d} \cdot \mathcal{L}_{diff} +  \lambda_{cont} \cdot \mathcal{L}_{cont} + \lambda_{ssim} \cdot \mathcal{L}_{ssim} + \lambda_{color} \cdot \mathcal{L}_{color}, \label{eq:lossII}
\end{equation}
where $\lambda_{d}$ is a hyper-parameter for balancing various losses.

\section{Experiments}
\subsection{Experimental Details}
\textbf{Implementation Details.}  Our restoration and fusion network inherits Restormer~\citep{Zamir2022Restormer}, a 4-level encoder-decoder Transformer architecture with degradation and semantic prior modulation. From level-1 to level-4, Transformer block numbers are set as $[2,2,4,4]$, and channel numbers are set as $[32, 64, 128, 256]$. SPEN contains $6$ residual blocks with token number and channel dimension set to $N_c=16$ and $C^\prime_c = 256$. DPEN contains $4$ residual blocks with token number and channel dimension set to $N_d=16$ and $C^\prime_d = 128$. The time-step of SPDM is set as $T=10$. We train our DSPFusion with the AdamW optimizer with $\beta_1=0.9$ and $\beta_2=0.99$. The learning rate is initialized to $2\times10^{-4}$ and gradually reduced to $1\times10^{-6}$ with cosine annealing. Both Stages \Rmnum{1} and \Rmnum{2} are trained for $50k$ iterations. The hyper-parameters are empirically set as $\gamma=0.75$, $\lambda_{cont}=15, \lambda_{ssim}=2, \lambda_{color}=20, \lambda_{cl}=1, \lambda_{d}=10$. The numbers of positive and negative samples are set to $K=3$ and $M=7$. Our training data is construed on the EMS dataset~\citep{Yi2024Text-IF}, including $2,210$ scenarios, with $8,804$ and $10,318$ low-quality infrared and visible images.

\textbf{Experiment Settings.} We first compare DSPFusion with nine SOTA fusion methods on four typical datasets (MSRS~\citep{Tang2022PIAFusion}, LLVIP~\citep{Jia2021LLVIP}, RoadScene~\citep{Xu2022U2Fusion}, TNO~\citep{Toet2017TNO}) with $4$ quantitative metrics, \emph{i.e.,} EN, MI, VIF, and $Q_{abf}$. Then, we validate its performance under various degradations, including blur, rain, low-light, over-exposure, and random noise in visible (VI), and low-contrast, random noise, and stripe noise in infrared~(IR). We also evaluate our robustness under mixed degradations, \emph{i.e.}, rain or low-light in VI, and low-contrast or stripe noise in IR. All scenarios include $100$ test samples, except for over-exposure in VI, containing $50$ test samples. Four no-reference metrics, \emph{i.e.}, MUSIQ, PI, TReS, and SD (or EN or SF), are used to verify fusion qualities. SOTA restoration algorithms are deployed to pre-enhance low-quality sources if fusion methods without enhancement for fair comparisons. Particularly, Hi-Diff~\citep{Chen2024Hi-Diff} for deblurring, NeRD-Rain~\citep{Chen2024NeRD-Rain} for deraining, Spadap~\citep{Li2023Spadap} for denoising, QuadPrior~\citep{Wang2024QuadPrior} for low-light enhancement, IAT~\citep{Cui2022IAT} for exposure correction, WDNN~\citep{Guan2019SNRWDNN} for stripe noise removal, and the method in \cite{Tang2022PIAFusion} for low-contrast enhancement.

\begin{figure*}[t]
	\centering
	\includegraphics[width=0.99\linewidth]{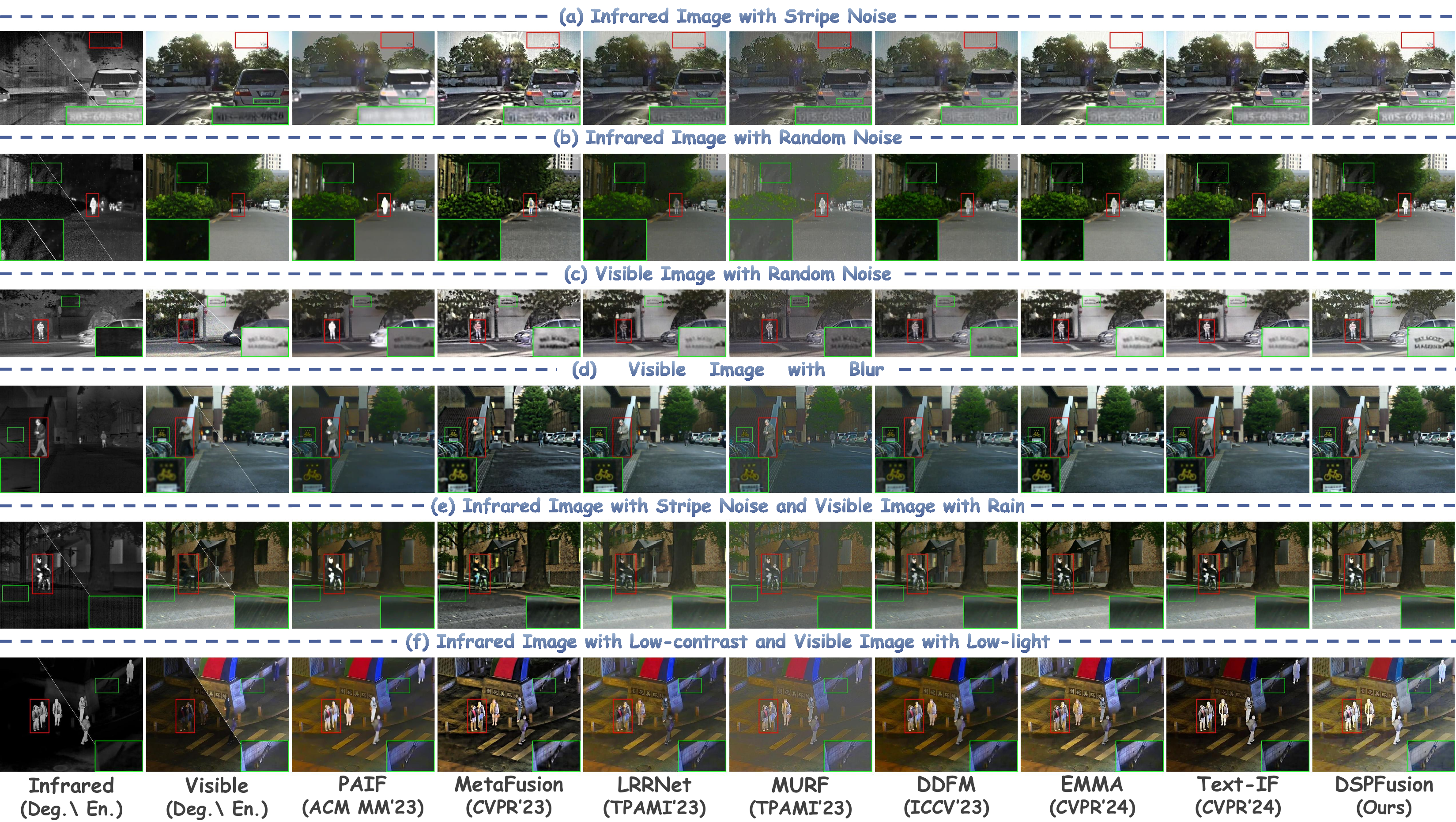}	
	\vspace{-0.05in}
	\caption{Visualization of fusion results in different degraded scenarios with pre-enhancement.}
	\label{fig:degradation}
\end{figure*}

\subsection{Fusion Performance Comparison}
\textbf{Comparison without Pre-Enhancement.}
Table~\ref{tab:normal} shows quantitative results on typical fusion datasets. DSPFusion achieves superior performance in MI and Qabf, effectively transferring complementary and edge information into fused images. The optimal VIF indicates that our fused images exhibit excellent visual perception quality, while the comparable EN suggests that our fusion results retain abundant information. In summary, the quantitative results demonstrate our remarkable fusion performance. Some visual fusion results are provided in the \textbf{Supple. Material}. 

\begin{figure}  
	\centering
	\vspace{-0.00in} 
	\includegraphics[width=1\linewidth]{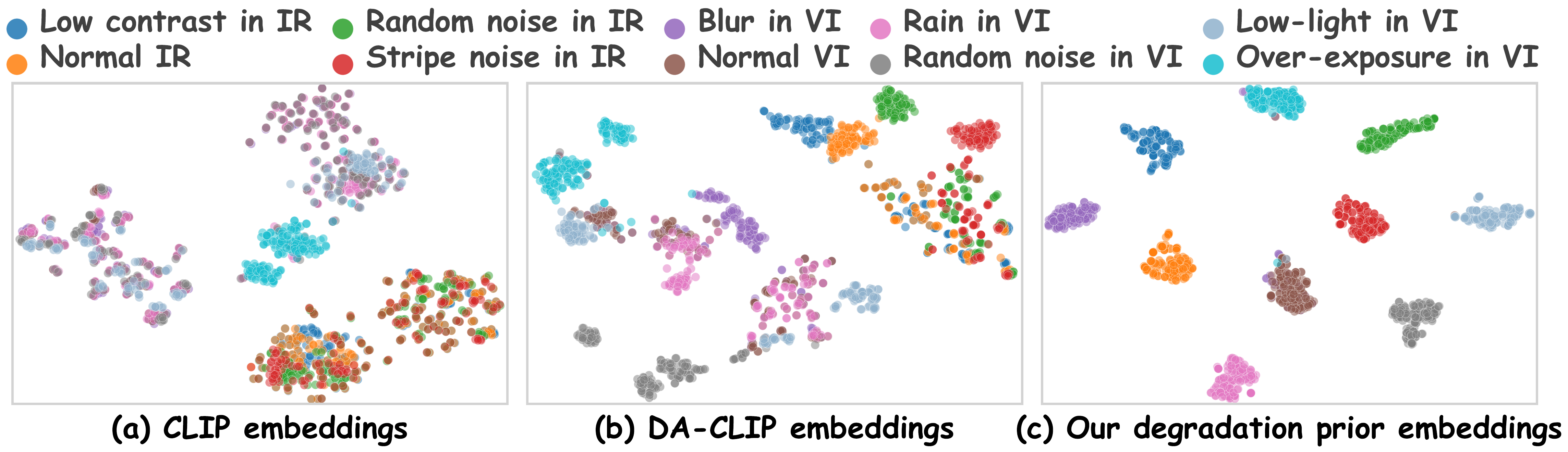}
	\vspace{-0.2in} 
	\caption{t-SNE plots of the degradation embeddings.}
	\label{fig:t-SNE}
\end{figure}

\textbf{Comparison with Pre-Enhancement.} The quantitative results in degraded scenarios are shown in Tab.~\ref{tab:degradation}. DSPFusion achieves the best MUSIQ, PI, and TReS in almost all degraded scenarios, demonstrating its effectiveness in mitigating degradation, aggregating complementary context, and producing high-quality fused images. We apply various no-reference statistical metrics, \emph{i.e.}, SD, EN, or SF, to evaluate fusion results on different degraded scenarios according to their properties. DSPFusion exhibits comparable performance to other methods on these metrics.

Qualitative results are presented in Fig.~\ref{fig:degradation}. When source images are affected by noise, denoising methods can remove noise but often at the cost of blurring fine details, such as texts on the wall and the license plate. By contrast, DSPFusion preserves rich texture while suppressing noise. In blurry scenarios, although HI-Diff partially mitigates the blur, DSPFusion delivers sharper visual clarity. This advantage arises from the fact that \textit{our method can enhance the degraded modality by leveraging comprehensive semantic priors from both infrared and visible images, offering a more complete scene representation}. Conversely, single-modality enhancement methods rely solely on limited intra-modality information to infer degradation-free images, which naturally limits their enhancement performance. As shown in Fig.~\ref{fig:degradation} (e) and (f), our method can effectively handle challenging scenarios where both infrared and visible images suffer from degradations. This is achieved by \textit{employing modality-specific degradation priors in a divide-and-conquer manner to modulate the features of each modality individually, ensuring that the feature enhancement is precisely adapted to the unique characteristics of each modality}. Both quantitative and qualitative results demonstrate the superiority of DSPFusion in suppressing degradations and integrating complementary context across various degraded scenarios in a unified model.

\begin{figure}
	\centering
	\vspace{-0.00in} 
	\includegraphics[width=1\linewidth]{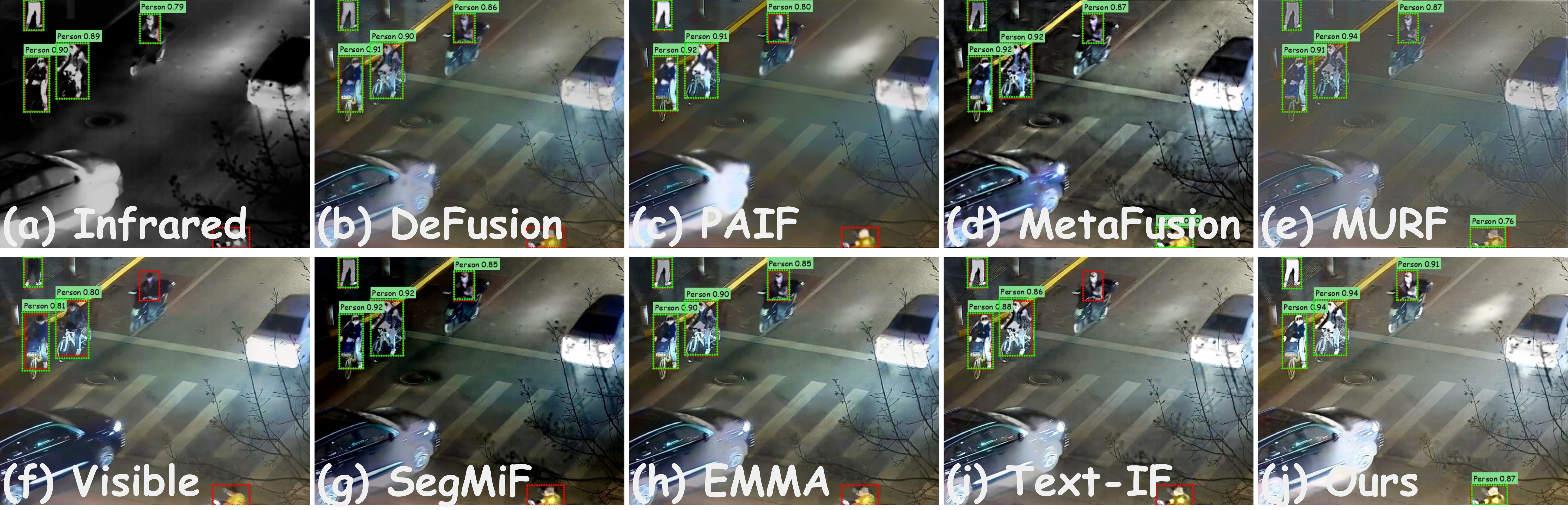}
	\vspace{-0.20in} 
	\caption{Visual comparison of object detection.}
	\vspace{-0.05in} 
	\label{fig:detection}
\end{figure}

\subsection{Extended Experiments and Discussions}
\textbf{Degradation Prior Visualization.} Figure~\ref{fig:t-SNE} shows t-SNE visualizations illustrating the ability of various models to distinguish degradation types. While DA-CLIP can partially separate degradations in visible images, it performs poorly on infrared images. In contrast, our DPEN effectively distinguishes degradations across modalities, laying a solid foundation for subsequent restoration and fusion.

\textbf{Object Detection.} We also evaluate object detection performance on LLVIP to indirectly assess the fusion quality with re-trained YOLOv8~\citep{Redmon2016YOLO}. Qualitative and quantitative results are shown in Fig.~\ref{fig:detection} and Tab.~\ref{tab:detection}. Owing to superior information restoration and integration, the detector identifies all pedestrians in our fusion results with higher confidence and achieves the best average precision (AP) across various confidence thresholds.

\begin{table}[t]
	\centering
	\caption{Quantitative results of object detection.} 	\label{tab:detection}	
	\vspace{-0.1in}
	\setlength{\tabcolsep}{4pt}
	\resizebox{0.99\linewidth}{!}{
		\begin{tabular}{@{}lcccccccccccl@{}}
			\toprule
			\multicolumn{1}{c}{} & \multicolumn{12}{l}{\textbf{Fusion in nighttime scenarios with enhancement}} \\ \cmidrule(l){2-13}
			\multicolumn{1}{c}{\multirow{-2}{*}{\textbf{Methods}}} & \textbf{Prec.} &  & \textbf{Recall} &  & \textbf{AP@50} & & \textbf{AP@75} &  & \textbf{AP@95} &  & \textbf{mAP} &  \\ \midrule
			\textbf{DeFus.}~\cite{Liang2022DeFusion} & 0.983 &  & 0.831 &  & 0.911 &  & 0.802 &  & 0.057 &  & 0.684 &  \\
			\textbf{PAIF}~\cite{Liu2023PAIF} & \cellcolor[HTML]{FFC7CE}\textbf{0.989} &  & 0.848 &  & 0.919 &  & 0.799 &  & 0.071 &  & 0.683 &  \\
			\textbf{MetaFus.}~\cite{Zhao2023MetaFusion} & 0.958 &  & 0.871 &  & 0.927 &  & 0.806 &  & \cellcolor[HTML]{E6E6FA}0.116 &  & 0.697 &  \\
			\textbf{LRRNet}~\cite{Li2023LRRNet} & \cellcolor[HTML]{FFC7CE}\textbf{0.989} &  & 0.811 &  & 0.902 &  & 0.783 &  & 0.040 &  & 0.668 &  \\
			\textbf{MURF}~\cite{Xu2023MURF} & 0.980 &  & \cellcolor[HTML]{FFC7CE}\textbf{0.884} &  & \cellcolor[HTML]{E6E6FA}0.935 &  & \cellcolor[HTML]{E6E6FA}0.809 &  & 0.095 &  & \cellcolor[HTML]{E6E6FA}0.707 &  \\
			\textbf{SegMiF}~\cite{Liu2023SegMiF} & 0.981 &  & 0.835 &  & 0.910 &  & 0.799 &  & 0.092 &  & 0.687 &  \\
			\textbf{DDFM}~\cite{Zhao2023DDFM} & 0.983 &  & 0.842 &  & 0.917 &  & 0.801 &  & 0.067 &  & 0.679 &  \\
			\textbf{EMMA}~\cite{Zhao2024EMMA} & 0.971 &  & 0.846 &  & 0.916 &  & 0.791 &  & 0.085 &  & 0.685 &  \\
			\textbf{Text-IF}~\cite{Yi2024Text-IF} & \cellcolor[HTML]{FFC7CE}\textbf{0.989} &  & 0.782 &  & 0.887 &  & 0.778 &  & 0.043 &  & 0.667 &  \\
			\textbf{DSPFusion} & 0.974 &  & \cellcolor[HTML]{FFC7CE}\textbf{0.884} &  & \cellcolor[HTML]{FFC7CE}\textbf{0.936} &  & \cellcolor[HTML]{FFC7CE}\textbf{0.822} &  & \cellcolor[HTML]{FFC7CE}\textbf{0.169} &  & \cellcolor[HTML]{FFC7CE}\textbf{0.726} &  \\ \bottomrule
		\end{tabular}
	}
\end{table}%

\begin{figure}
	\centering
		\vspace{-0.05in} 
	\includegraphics[width=0.97\linewidth]{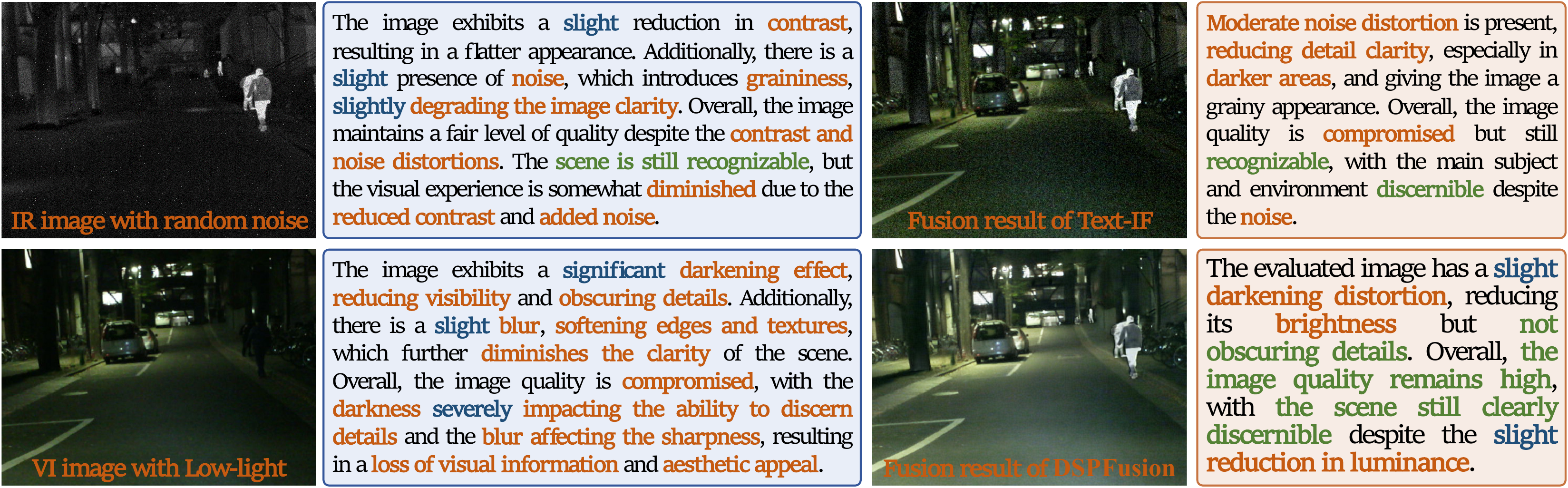}
	\vspace{-0.05in} 
	\caption{Evaluation results with DepictQA.}
		\vspace{-0.05in} 
	\label{fig:DepictQA}
\end{figure}

\textbf{Evaluation with DepictQA.} We introduce DepictQA~\citep{You2024DepictQA}, a descriptive image quality assessment metric based on vision language models, to evaluate our fused image quality. As shown in Fig.~\ref{fig:DepictQA}, the infrared image suffers from significant noise, while the visible image is affected by low-light. DepictQA not only accurately identifies these degradations but also describes their severity. Although Text-IF only mildly suppresses degradations, thanks to effective information aggregation, DepictQA judges that while its fusion result experiences moderate noise distortion, it remains recognizable. In contrast, DSPFusion successfully achieves low-light enhancement and noise reduction, along with effective information aggregation. Thus, DepictQA assesses our image quality as remaining high.

\textbf{Computational Efficiency.} We conduct the diffusion process in a compact space, greatly reducing computational costs. As shown in Tab.~\ref{tab:efficiency}, compared to DDFM, which performs diffusion in the image space, DSPFusion exhibits a significant advantage, being over $29$ faster than DDFM. Moreover, in comparison to Text-IF, which relies on an additional CLIP model for degradation prompting, DSPFusion also offers a notable improvement in efficiency. Specifically, in degraded scenarios, it offers a clear advantage by obviating the need for additional pre-processing.

\begin{table}[t]
	\centering
	\caption{Computational efficiency comparison on MSRS dataset.}	\label{tab:efficiency}
	\vspace{-0.1in}
	\setlength{\tabcolsep}{3pt}
	\resizebox{0.99\linewidth}{!}{
		\begin{tabular}{@{}lclclclcclclcl@{}}
			\toprule
			\multicolumn{1}{c}{} & \multicolumn{6}{l}{\textbf{Fusion}} & \multicolumn{1}{l}{} & \multicolumn{6}{l}{\textbf{Fusion with enhancement}} \\ \cmidrule(lr){2-7} \cmidrule(l){9-14}
			\multicolumn{1}{l}{\multirow{-2}{*}{\textbf{Methods}}} & \textbf{parm.(m)} & \multicolumn{1}{c}{\textbf{}} & \textbf{flops(g)} & \multicolumn{1}{c}{\textbf{}} & \textbf{time(s)} & \multicolumn{1}{c}{\textbf{}} & \textbf{} & \textbf{parm.(m)} & \multicolumn{1}{c}{\textbf{}} & \textbf{flops(g)} & \multicolumn{1}{c}{\textbf{}} & \textbf{time(s)} &  \\ \midrule
			\textbf{DeFus.}~\cite{Liang2022DeFusion} & 7.874 &  & 71.55 &  & 0.075 &  &  & 234.81 &  & 869.24 &  & 0.478 &  \\
			\textbf{PAIF}~\cite{Liu2023PAIF} & 44.86 &  & 122.12 &  & 0.052 &  &  & 271.80 &  & 919.80 &  & 0.455 &  \\
			\textbf{MetaFus.}~\cite{Zhao2023MetaFusion} & 0.812 &  & 159.48 &  & \cellcolor[HTML]{FFC7CE}\textbf{0.028} &  &  & 227.74 &  & 957.16 &  & 0.431 &  \\
			\textbf{LRRNet}~\cite{Li2023LRRNet} & \cellcolor[HTML]{FFC7CE}\textbf{0.049} &  & \cellcolor[HTML]{FFC7CE}\textbf{14.17} &  & 0.085 &  &  & 226.98 &  & \cellcolor[HTML]{E6E6FA}811.86 &  & 0.488 &  \\
			\textbf{MURF}~\cite{Xu2023MURF} & \cellcolor[HTML]{E6E6FA}0.120 &  & \cellcolor[HTML]{E6E6FA}31.50 &  & 0.205 &  &  & 227.05 &  & 829.19 &  & 0.608 &  \\
			\textbf{SegMiF}~\cite{Liu2023SegMiF} & 45.04 &  & 353.7 &  & 0.147 &  &  & 271.97 &  & 1151.4 &  & 0.550 &  \\
			\textbf{DDFM}~\cite{Zhao2023DDFM} & 552.7 &  & 5220. &  & 3.451 &  &  & 779.59 &  & 6018.2 &  & 3.861 &  \\
			\textbf{EMMA}~\cite{Zhao2024EMMA} & 1.516 &  & 41.54 &  & \cellcolor[HTML]{E6E6FA}0.037 &  &  & 228.45 &  & 839.23 &  & 0.440 &  \\
			\textbf{Text-IF}~\cite{Yi2024Text-IF} & 89.01 &  & 1518.9 &  & 0.157 &  &  & \cellcolor[HTML]{E6E6FA}89.01 &  & 1518.9 &  & \cellcolor[HTML]{E6E6FA}0.157 &  \\
			\textbf{DSPFusion} & 13.99 &  & 254.34 &  & 0.119 &  &  & \cellcolor[HTML]{FFC7CE}\textbf{13.99} &  & \cellcolor[HTML]{FFC7CE}\textbf{254.34} &  & \cellcolor[HTML]{FFC7CE}\textbf{0.119} &  \\
			\bottomrule
		\end{tabular}
	}
\end{table}

\begin{figure} 
	\centering
	\includegraphics[width=1\linewidth]{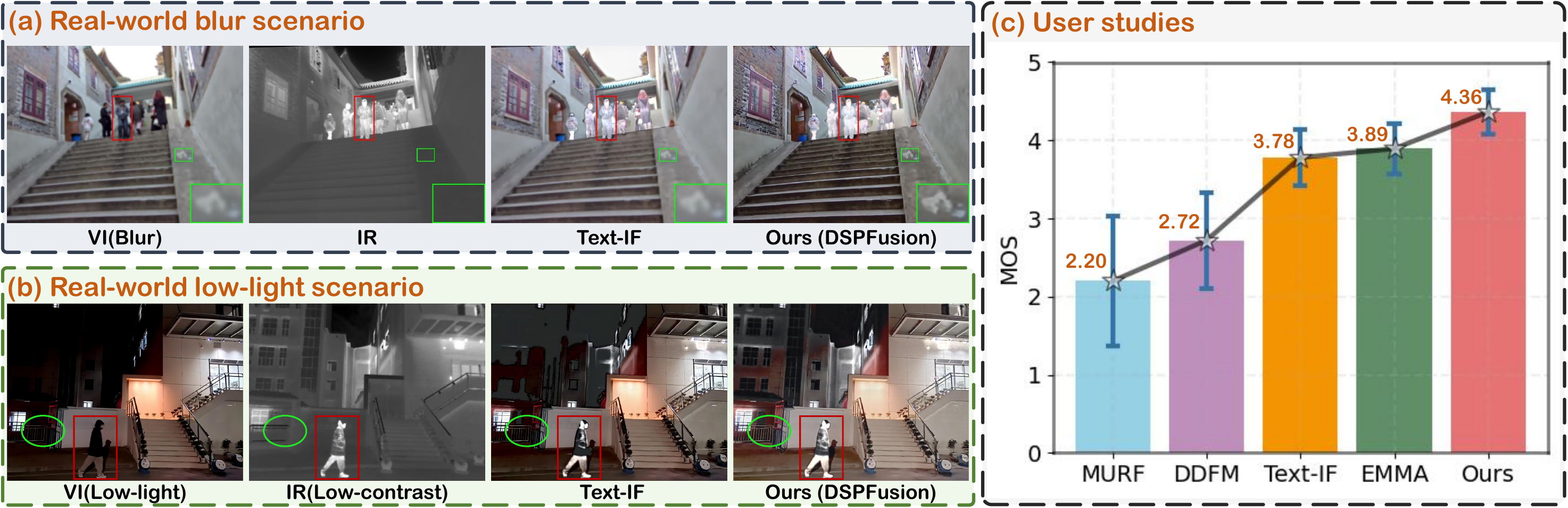}
	\vspace{-0.20in} 
	\caption{Generalization results in the real world and user studies.}
	\vspace{-0.05in} 
	\label{fig:user_study}
\end{figure}

\textbf{Real-World Generalization.} As shown in Fig.~\ref{fig:user_study}~(a) and (b), our method generalizes well to real-world scenarios. In the blur scenario, although the visible image suffers from blur degradation due to focal plane vibration, DSPFusion effectively restores textural details. Besides, in the low-light condition, our fusion result significantly enhances scene visibility while maintaining natural contrast.

\textbf{User Studies.} $30$ participants are invited to score~fusion results of $50$ image pairs of various degradation settings or real-world scenarios. As shown in Fig.~\ref{fig:user_study}~(c), our method achieves the highest MOS, indicating that our results are more in line with user perception.

\begin{table} 
	\centering
	\caption{Quantitative results of the ablation studies.} \label{tab:ablation}
	\vspace{-0.08in} 
	\setlength{\tabcolsep}{1pt}
	\resizebox{0.99\linewidth}{!}{
		\begin{tabular}{@{}ccccccccccclcccccccccccccccccccccccccc@{}}
			\toprule
			&  &  & \multicolumn{8}{l}{\textbf{VI(Blur)}} &  & \multicolumn{8}{l}{\textbf{VI(Rain)}} & \multicolumn{1}{l}{\textbf{}} & \multicolumn{8}{l}{\textbf{IR(Stripe noise, SN)}} & \multicolumn{1}{l}{\textbf{}} & \multicolumn{8}{l}{\textbf{VI(Rain) and IR(SN)}} \\ \cmidrule(lr){4-11} \cmidrule(lr){13-20} \cmidrule(lr){22-29} \cmidrule(l){31-38}
			\multirow{-2}{*}{\textbf{Configs}} & \multirow{-2}{*}{\textbf{\begin{tabular}[c]{@{}c@{}}Deg. \\ prior\end{tabular}}} & \multirow{-2}{*}{\textbf{\begin{tabular}[c]{@{}c@{}}Sema. \\ prior\end{tabular}}} & \textbf{MUSIQ} & \textbf{} & \textbf{PI} & \textbf{} & \textbf{Tres} & \textbf{} & \textbf{SF} & \textbf{} &  & \textbf{MUSIQ} & \textbf{} & \textbf{PI} & \textbf{} & \textbf{Tres} & \textbf{} & \textbf{SD} & \textbf{} & \textbf{} & \textbf{MUSIQ} & \textbf{} & \textbf{PI} & \textbf{} & \textbf{Tres} & \textbf{} & \textbf{EN} & \textbf{} & \textbf{} & \textbf{MUSIQ} & \textbf{} & \textbf{PI} & \textbf{} & \textbf{Tres} & \textbf{} & \textbf{EN} & \textbf{} \\ \midrule
			\textbf{I} & \XSolidBrush & \XSolidBrush & 45.88 &  & 3.21 &  & 46.49 &  & 15.10 &  &  & 49.07 &  & 2.69 &  & 55.03 &  & 54.27 &  &  & 50.22 &  & 2.89 &  & 56.137 &  & 7.17 &  &  & 50.06 &  & 2.85 &  & 54.89 &  & \cellcolor[HTML]{E6E6FA}7.26 &  \\
			\textbf{II} & \Checkmark & \XSolidBrush & \cellcolor[HTML]{E6E6FA}46.96 &  & \cellcolor[HTML]{E6E6FA}3.07 &  & \cellcolor[HTML]{E6E6FA}48.44 &  & \cellcolor[HTML]{E6E6FA}15.46 &  &  & 49.61 &  & 2.68 &  & 55.92 &  & \cellcolor[HTML]{E6E6FA}55.20 &  &  & 50.80 &  & 2.80 &  & 56.42 &  & \cellcolor[HTML]{E6E6FA}7.27 &  &  & 50.52 &  & 2.79 &  & 55.96 &  & 7.17 &  \\
			\textbf{III} & \XSolidBrush & \Checkmark & 46.45 &  & 3.09 &  & 48.31 &  & 15.38 &  &  & \cellcolor[HTML]{E6E6FA}49.71 &  & \cellcolor[HTML]{FFC7CE}\textbf{2.53} &  & \cellcolor[HTML]{E6E6FA}56.19 &  & 54.83 &  &  & \cellcolor[HTML]{E6E6FA}50.85 &  & \cellcolor[HTML]{FFC7CE}\textbf{2.75} &  & \cellcolor[HTML]{E6E6FA}56.74 &  & 7.26 &  &  & \cellcolor[HTML]{E6E6FA}50.60 &  & \cellcolor[HTML]{E6E6FA}2.76 &  & \cellcolor[HTML]{E6E6FA}56.87 &  & 7.16 &  \\
			\textbf{Ours} & \Checkmark & \Checkmark & \cellcolor[HTML]{FFC7CE}\textbf{47.14} &  & \cellcolor[HTML]{FFC7CE}\textbf{2.97} &  & \cellcolor[HTML]{FFC7CE}\textbf{49.75} &  & \cellcolor[HTML]{FFC7CE}\textbf{15.69} &  &  & \cellcolor[HTML]{FFC7CE}\textbf{50.47} &  & \cellcolor[HTML]{E6E6FA}2.56 &  & \cellcolor[HTML]{FFC7CE}\textbf{56.53} &  & \cellcolor[HTML]{FFC7CE}\textbf{55.60} &  &  & \cellcolor[HTML]{FFC7CE}\textbf{51.06} &  & \cellcolor[HTML]{E6E6FA}2.79 &  & \cellcolor[HTML]{FFC7CE}\textbf{57.24} &  & \cellcolor[HTML]{FFC7CE}\textbf{7.35} &  &  & \cellcolor[HTML]{FFC7CE}\textbf{50.88} &  & \cellcolor[HTML]{FFC7CE}\textbf{2.74} &  & \cellcolor[HTML]{FFC7CE}\textbf{57.10} &  & \cellcolor[HTML]{FFC7CE}\textbf{7.35} &  \\ \bottomrule
		\end{tabular}
	}
	\vspace{-0.22in} 
\end{table}

\textbf{Ablation Studies.} In order to demonstrate the effectiveness of our specific designs, we conduct ablation studies by individually removing either DPEN or SPEN in various degraded scenarios. From Tab.~\ref{tab:ablation}, one can find that both DPEN and SPEN play crucial roles in improving the performance of DSPFusion. Particularly, our method better reconciles degradation suppression with information aggregation by integrating degradation and semantic priors.

\section{Conclusion}
This work presents a degradation and semantic dual-prior guided framework for degraded image fusion. A degradation prior embedding network is designed to extract modality-specific degradation priors, guiding the unified model to purposefully address degradations. A semantic prior embedding network is developed to capture semantic prior from cascaded source images, enabling implicit complementary information aggregation. Moreover, we devise a semantic prior diffusion model to restore high-quality scene priors in a compact space, providing global semantic guidance for subsequent restoration and fusion. Experiments on multiple degraded scenarios demonstrate our superiority in suppressing degradation and aggregating information.

{
    \small
    \bibliographystyle{ieeenat_fullname}
    \bibliography{main}
}

\maketitlesupplementary
\renewcommand\thesection{\Alph{section}}
\setcounter{section}{0} 
\section{More Details About Methodology Designs}
\begin{figure}[t]
	\centering
	\includegraphics[width=0.98\linewidth]{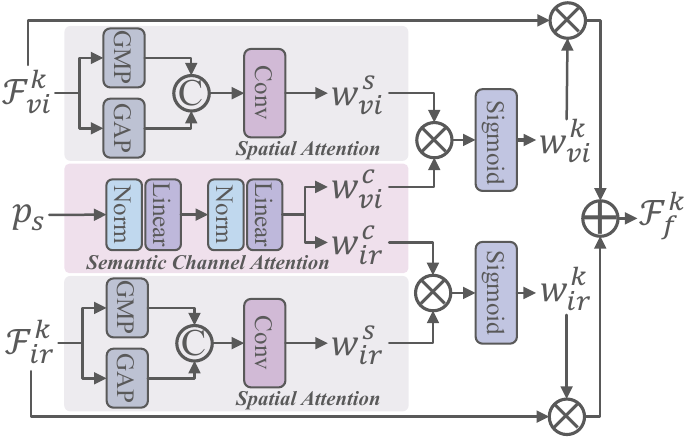}
	\caption{The architecture of the PGFM.}
	\label{fig:PGFM}
\end{figure}

In this section, we provide more details about our methodology designs. Figure~\ref{fig:PGFM} presents the network architecture of our prior-guided fusion module~(PGFM). Given the global semantic prior~$p_s$, which integrates high-quality and comprehensive scene contexts, we utilize a semantic channel attention, consisting of two linear layers, to generate a channel-wise fusion weight~($w_{ir}^c$ or $w_{vi}^c$). Additionally, a spatial attention is employed to achieve spatial activity level measurements. The infrared (or visible) features are compressed along the channel dimension via global max pooling~(GMP) and global average pooling~(GAP). The pooled results are then concatenated along the channel dimension and fed into a convolutional layer to generate spatial weights~($w_{ir}^s$ or $w_{vi}^s$). Subsequently, we comprehensively integrate the channel- and spatial-wise attention to obtain the final fusion weight, formulated as:
\begin{equation}
	w_{ir}= \sigma(w_{ir}^c \otimes w_{ir}^s), \quad  \quad w_{vi} = \sigma(w_{vi}^c \otimes w_{vi}^s),
\end{equation}
where $\otimes$ denotes element-wise multiplication with broadcasting, and $\sigma$ is the sigmoid function. Finally, the fusion process is defined as:
\begin{equation}
	\mathcal{F}_f = w_{ir}  \mathcal{F}_{ir} \oplus w_{vi} \mathcal{F}_{vi}.
\end{equation}

Additionally, unlike the encoder layer, the prior-modulated decoder layer employs the semantic prior integration module instead of the dual-prior guidance module, relying solely on high-quality scene semantic priors to support feature reinforcement. This design aims to effectively eliminate the influence of degradation factors during the feature encoding stage with the assistance of degradation and semantic priors. After feature fusion, if fused features still contain mixed degradations, their distribution will differ from that of a single degradation, making it challenging for the degradation prior to characterize them accurately. 

\begin{figure}
	\centering
	\includegraphics[width=1\linewidth]{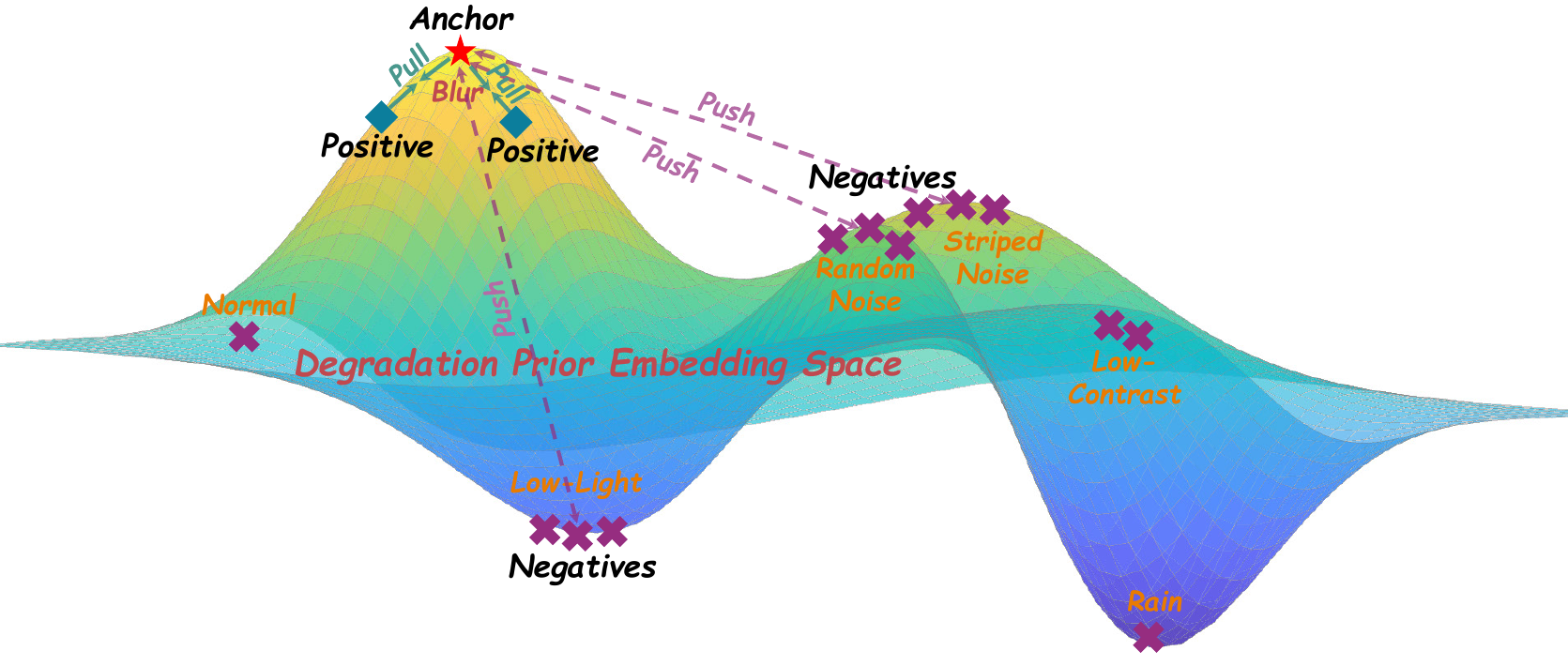}	
	\caption{Schematic diagram of the contrastive mechanism.}
	\label{fig:CL}
\end{figure}

\begin{figure*}[t]
	\centering
	\includegraphics[width=0.99\linewidth]{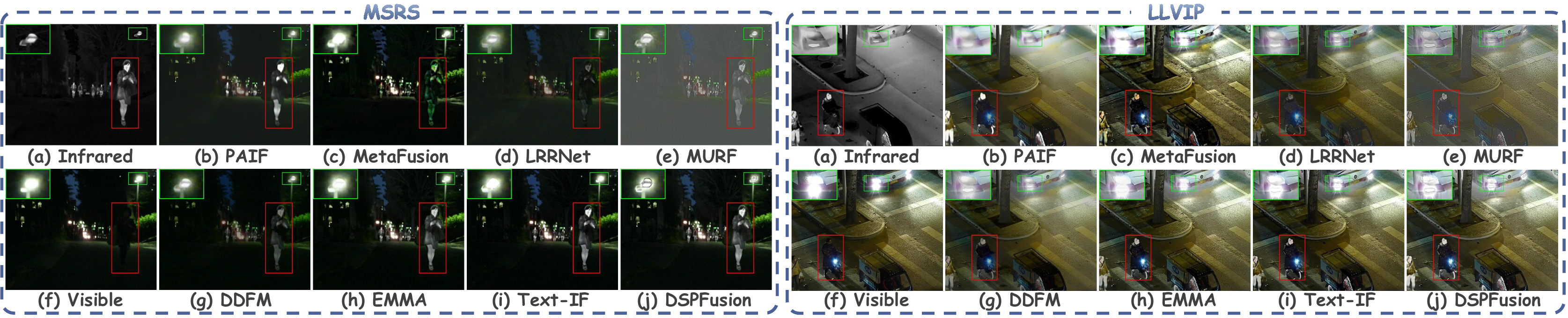}	
	\caption{Visualization of fusion results on the typical fusion datasets.}
	\label{fig:norm}
\end{figure*}

\begin{figure*}[t]
	\centering
	\includegraphics[width=0.99\linewidth]{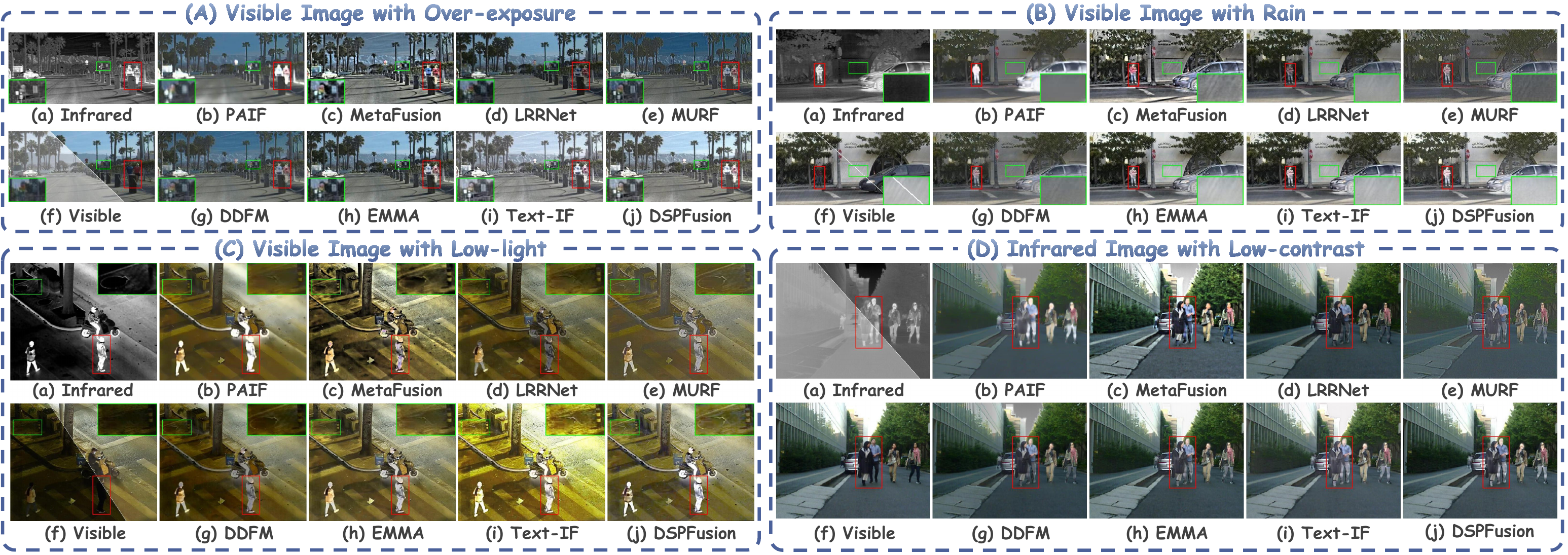}	
	\caption{Visualization of fusion results in degradation scenarios with pre-enhancement.}
	\label{fig:supplementary}
\end{figure*}

Figure~\ref{fig:CL} presents the schematic diagram of our contrastive mechanism for constraining the degradation prior embedding network. This section focuses on the criteria for selecting the positive and negative samples. The number of positive samples, $K$, is set to $3$, and the number of negative samples, $M$, is set to $7$. For instance, the degradation types of the anchors in visible and infrared images are low-light and low-contrast, respectively. For visible images, the positive samples consist of $3$ low-light visible images from different scenes, while for infrared images, the positive samples are $3$ low-contrast infrared images from different scenes. We then select $6$ negative samples for both visible and infrared anchors from the remaining visible or infrared images with the same scene content as the anchors, where the visible images suffer from over-exposure, blur, rain, random noise, or no degradation, while the infrared images are affected by random noise, stripe noise, or no degradation. Moreover, the infrared anchor is added as the negative sample for the visible anchor, and vice versa. Therefore, for each anchor, there are $3$ positive samples and $7$ negative samples.

\section{More Experiment Details}
\subsection{Implementation Details}
We construct the training data on the EMS dataset~\footnote{https://github.com/XunpengYi/EMS}, where the degradation types for visible images include blur, rain, low-light, over-exposure, and random noise, and the degradations for infrared images include low-contrast, random noise, and stripe noise. In particular, VI(Blur) uses Gaussian blur (kernel size $=21$, $\sigma=2.6$) to attenuate high-frequency details. VI(Rain) simulates rain stripes. VI(RN) and IR(RN) are influenced by Gaussian noise ($\mu=0$, $\sigma=10$) and Poisson noise ($\mu\in$[70, 90]). IR(SN) combines Gaussian noise ($\mu=0$, $\sigma\in[10, 15]$) and stripe noise ($\mu\in[20, 30]$). We further extend this dataset by introducing low-light scenes from the MSRS dataset, where the visible images are enhanced by QuadPrior~\cite{Wang2024QuadPrior}. Therefore, VI(LL) from MSRS and LLVIP, VI(OE) from RoadScene, and IR(LC) from MSRS. Finally, our training dataset consists of $2,210$ paired high-quality infrared and visible images. The source infrared images include $2,210$ degradation-free images, $2,210$ low-contrast images, $2,210$ images with random noise, and $2,210$ images with stripe noise. The source visible images involve $2,210$ degradation-free images, $2,210$ blurred images, $2,210$ rain-affected images, $2,210$ images with random noise, $1,316$ low-light images, and $136$ over-exposed images. The contrastive learning configuration follows AirNet~\cite{Li2022AirNet} and sampling steps refer to Hi-Diff~\cite{Chen2024Hi-Diff}. All experiments are conducted on the NVIDIA RTX 4090 GPUs and 2.50 GHz Intel(R) Xeon(R) Platinum 8180 CPUs with PyTorch.

\begin{figure*}[t]
	\centering
	\includegraphics[width=1\linewidth]{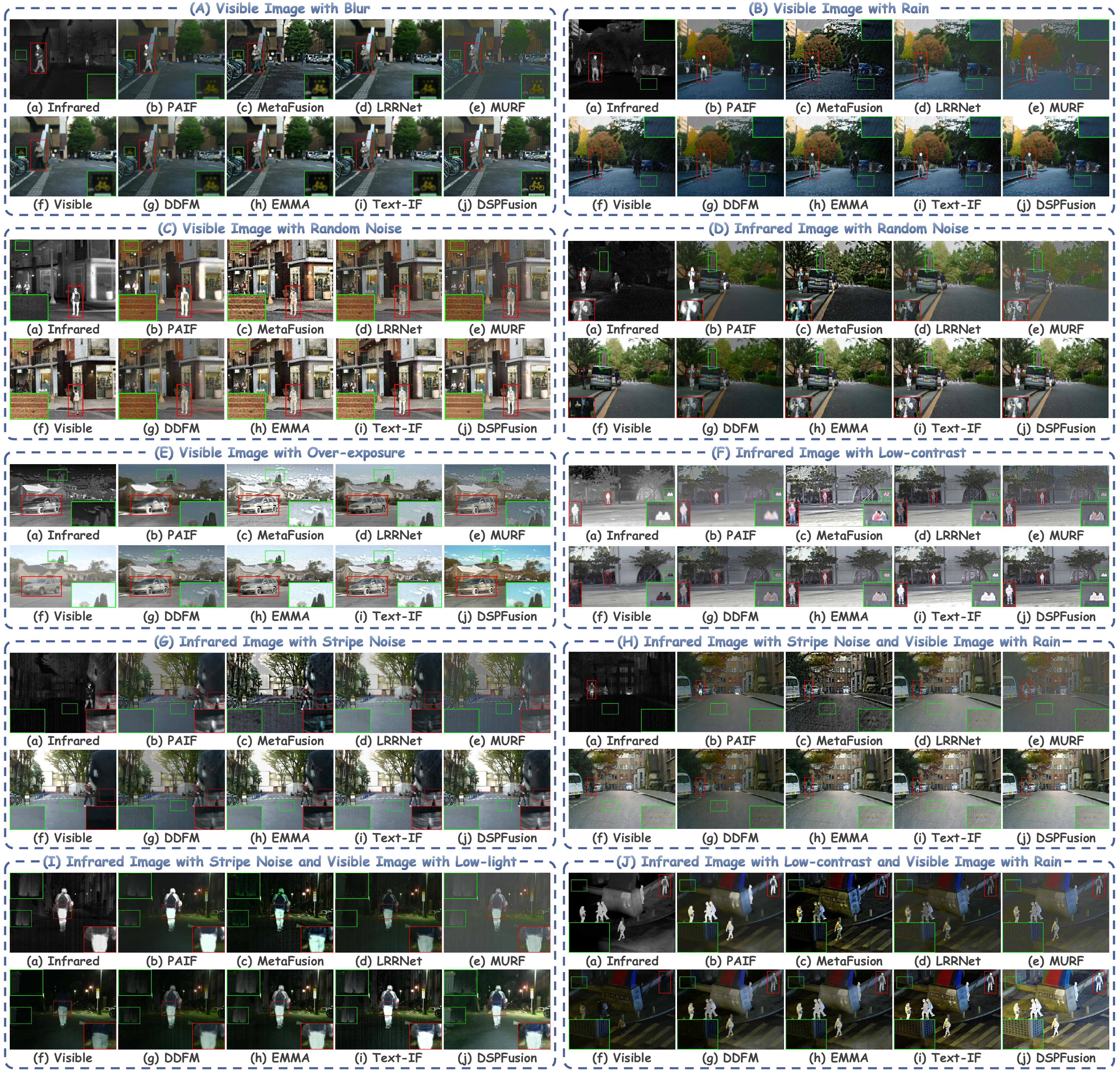}	
	\caption{Visualization of fusion results in degradation scenarios without pre-enhancement.}
	\label{fig:no-enhancemnt}
\end{figure*}

\subsection{Experiment Configures}
The nine state-of-the-art fusion methods includes DeFusion~\cite{Liang2022DeFusion}, PAIF~\cite{Liu2023PAIF}, MetaFusion~\cite{Zhao2023MetaFusion}, LRRNet~\cite{Li2023LRRNet}, MURF~\cite{Xu2023MURF}, SegMiF~\cite{Liu2023SegMiF}, DDFM~\cite{Zhao2023DDFM}, EMMA~\cite{Zhao2024EMMA}, Text-IF~\cite{Yi2024Text-IF}. The numbers of test images in the MSRS, LLVIP, RoadScene, and TNO datasets are $361$, $50$, $50$, and $25$, respectively, for quantitative fusion comparison without enhancement. When quantitatively evaluating performance under degradation scenarios, all scenarios include $100$ test samples, except for the over-exposed scenario in visible images, which contains $50$ test samples. For a fair comparison, almost all fusion algorithms apply state-of-the-art image restoration methods to pre-enhance source images. Notably, Text-IF~\cite{Yi2024Text-IF} utilizes its built-in enhancement module for low-light and over-exposed visible images, as well as low-contrast and random noise in infrared images, while applying pre-processing algorithms for other degradation scenarios. Additionally, when source images are affected by random noise, PAIF~\cite{Liu2023PAIF} does not employ additional denoising algorithms for pre-enhancement, as its fusion network is inherently robust to noise. 

\section{More Results and Analysis} 
Figure~\ref{fig:norm} presents representative visual fusion results on the MSRS and LLVIP datasets. We can find that MetaFusion, LRRNet, MURF, and DDFM diminish the prominence of thermal targets, while PAIF, EMMA, and Text-IF struggle to outline streetlights and headlights in overexposed conditions. In contrast, DSPFusion simultaneously highlights significant targets and preserves abundant textures. Overall, the quantitative and qualitative results collectively demonstrate the impressive fusion performance of our DSPFusion.

\begin{table*}[t]
	\centering
	\caption{Computational efficiency of pre-enhancement algorithms.} \label{tab:pre-efficiency}
	\setlength{\tabcolsep}{4pt}
	\resizebox{0.99\textwidth}{!}{
		\begin{tabular}{@{}lccccccc@{}}
			\toprule
			\textbf{Task} & Deblurring & Deraining & Denoising & \begin{tabular}[c]{@{}c@{}}Low-light\\ enhancement\end{tabular} & \begin{tabular}[c]{@{}c@{}}Exposure \\ correction\end{tabular} & \begin{tabular}[c]{@{}c@{}}Stripe noise\\ remove\end{tabular} & \multirow{2}{*}{Average} \\ \cmidrule(r){1-7}
			\textbf{Method} & Hi-Diff & NeRD-Rain & Spadap & QuadPrior & IAT & WDNN &  \\ \midrule
			\textbf{Parm. (M)} & 24.152 & 22.856 & 1.084 & 1313.39 & 0.087 & 0.013 & 226.93 \\
			\textbf{Flops (G)} & 529.359 & 693.649 & 81.875 & 3473.413 & 6.728 & 1.105 & 797.688 \\
			\textbf{Time (s)} & 0.359 & 0.299 & 0.003 & 1.745 & 0.007 & 0.003 & 0.403 \\ \bottomrule
		\end{tabular}
	}
\end{table*}

\begin{figure*}[t]
	\centering
	\includegraphics[width=1\linewidth]{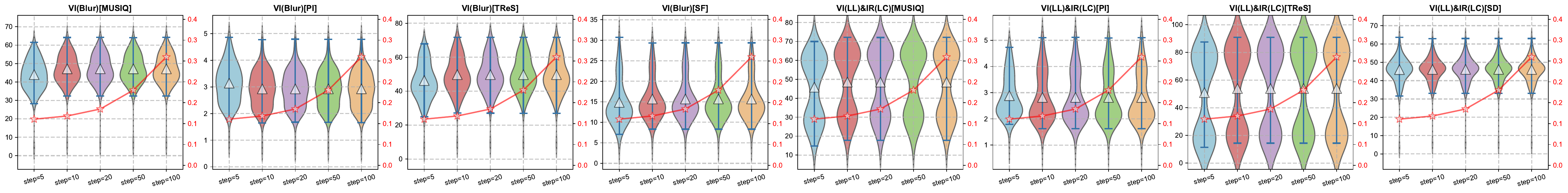}
	\caption{Quantitative analysis of various sampling steps.}\label{fig:steps}
\end{figure*}

In the degradation scenarios, besides using no-reference image quality assessment metrics, \emph{i.e.,} MUSIQ, PI, and TReS, we also utilize statistical metrics frequently employed in the image fusion field to assess performance based on the properties of degradations. In detail, when source images are affected by blurring, textures become obscured. Therefore, we use the SF metric to evaluate the richness of details in the fusion results. Additionally, when source images suffer from issues such as low light, overexposure, or low contrast, the overall contrast diminishes. Consequently, we use the SD metric to assess the effectiveness of the fusion results in counteracting these degradations. Furthermore, when images are affected by noise or rain, both SF and SD values may be artificially inflated, so we employ the EN metric to evaluate the fusion performance accurately.

Figure~\ref{fig:supplementary} provides more fusion results in the degradation scenarios with enhancement. From Fig~\ref{fig:supplementary}, one can find that PAIF obscures texture details within the scenes, particularly in prominent infrared targets, despite the excessive enhancement of these targets. This is attributed to PAIF attempting to counteract noise. Additionally, MetaFusion introduces artificial textures during the fusion process, which is the primary factor for its higher SF metric. We believe this is caused by MetaFusion paying more attention to the object detection task, resulting in insufficient consideration for visual perception. LRRNet, MURF, and DDFM seem to simply neutralize infrared and visible images, resulting in their fusion results that reduce the prominence of infrared targets and cause a loss of texture details in the visible images. EMMA relies on manually selected fused images from existing fusion algorithms for supervision, which limits its performance potential. For instance, while EMMA can aggregate complementary information from source images across most scenarios, it may slightly diminish the prominence of infrared targets. Although Text-IF demonstrates good fusion performance, it still has several notable shortcomings. Firstly, Text-IF is highly sensitive to text prompts. As shown in our main manuscript, when we prompt it that the visible and infrared images suffer from degradations (such as low light and low contrast) simultaneously, it fails to mitigate the effects of degradations, even though it handles individual degradations effectively, as demonstrated in Fig.~\ref{fig:supplementary}~(C) and (D). This may be caused by the fact that the feature embedding of such coupled text prompts is unfamiliar to the pre-trained model. Additionally, it is limited to addressing only a few specific types of degradations, such as low-light and over-exposure in visible images as well as random noise and low-contrast in infrared images. In contrast, our method adaptively identifies degradation types from the source images, enabling it to effectively handle the common degradations and achieve complementary information aggregation. Moreover, by employing a divide-and-conquer manner to address degradations in infrared and visible images separately, it remains effective even when both infrared and visible images suffer from degradations simultaneously.


Figure~\ref{fig:no-enhancemnt} presents the qualitative comparison results in degradation scenarios without pre-enhancement. It is evident that although most fusion algorithms can effectively aggregate complementary information, they are hindered by degradations and cannot provide satisfactory fusion outcomes. PAIF is capable of handling noise-related degradations, but it tends to blur the structures and details in the scene, resulting in suboptimal results. Text-IF can address illumination degradation in visible images, as well as low contrast and random noise in infrared images, but it is ineffective against other common degradations. In contrast, our DSPFusion is able to consistently synthesize impressive fusion results across all degradation conditions. This is attributed to the fact that our degradation prior embedding network can adaptively identify degradation types, and semantic prior diffusion model effectively recovers high-quality semantic priors. The degradation priors and high-quality semantic priors complement each other, jointly guiding the restoration and fusion model.
%

\textbf{Effects of Sampling Steps.} The impact of different sampling steps on fusion performance is illustrated in Fig.~\ref{fig:steps}. The results reveal a nearly linear relationship between inference time and the number of sampling steps. Notably, when the number of sampling steps exceeds 10, further increases do not lead to a significant improvement in fusion performance. Consequently, we set the number of sampling steps to 10 to strike a balance between fusion quality and sampling time, which is also the default setting in Hi-Diff~\cite{Chen2024Hi-Diff}.

\textbf{Computational Efficiency of Pre-enhancement.} Table~\ref{tab:pre-efficiency} illustrates the computational efficiency of different pre-enhancement algorithms. From the results, we can find that some pre-enhancement algorithms, such as Spadap, IAT, and WDNN, are computationally efficient, while others, like Hi-Diff, NeRD-Rain, and QuadPrior, introduce heavy computational burdens. In particular, QuadPrior incurs significant computational costs as it conducts the diffusion process in the image domain. Our semantic prior diffusion model recovers high-quality semantic priors in a compact latent space, which greatly conserves computational overhead. We employ task-specific SOTA image enhancement methods for pre-enhancement, rather than relying on general approaches. On the one hand, general methods cannot simultaneously handle degradations in both infrared and visible modalities. On the other hand, general methods exhibit poor generalization on the infrared and visible image fusion datasets.

\end{document}